\documentclass[times, onecolumn]{zHenriquesLab-StyleBioRxiv}
\usepackage{blindtext}

\leadauthor{He} 

\begin{document}

\title{Linguistic boundaries emerge from pressure-constrained processing dynamics}

\author[a,\Letter]{Rui He}
\author[b,c]{Philipp Homan}
\author[a,d]{Wolfram Hinzen}

\affil[a]{Grammar and Cognition Lab, Department of Translation \& Language Sciences, Universitat Pompeu Fabra, Barcelona 08018, Spain}
\affil[b]{Department of Adult Psychiatry and Psychotherapy, University Hospital of Psychiatry Zurich, University of Zurich, Zurich 8032, Switzerland}
\affil[c]{Neuroscience Center Zurich, University of Zurich and ETH Zurich, Zurich 8057, Switzerland}
\affil[d]{Institució Catalana de Recerca i Estudis Avançats, Barcelona 08010, Spain}

\maketitle

\begin{abstract}
\noindent
Human cognition organizes continuous experience into bounded units while maintaining continuity across them. Linguistic elements are also organized into bounded structural units such as words and phrases, with the sentence marking the final bound of a grammatical construction process. How do such boundaries emerge? We propose that boundaries emerge from interacting demands to maintain unresolved dependencies, integrate completed relations, and update semantic context, with structural completion allowing but not determining closure. First, in an abstract generative simulation without predefined boundary positions or constituent labels, these demands produced sentence-like segment-length distributions and distinct characteristic timescales, with structural completion allowing but not determining closure. Near boundaries, unresolved dependency load declined, integration cost peaked, and accumulated pressure reset. Then, we tested whether the same demands predicted human listeners’ boundary choices during comprehension. Listeners were more likely to mark boundaries with fewer unresolved dependencies, higher integration cost and semantic mismatch, and elevated preceding memory load, beyond syntactic and prosodic cues. Third, in naturalistic self-paced reading, sentence-final slowing coincided with declining memory load and increasing integration cost and semantic mismatch, broadly consistent with the simulation.  Processing the same narratives with a large language model revealed increased local intrinsic dimensionality and straighter hidden-state trajectories at sentence endings, followed by larger representational updates. Explicit boundary status explained additional variation in reading time and hidden-state geometry beyond graded processing demands, but considerably less in next-token surprisal. Together, these findings suggest that linguistic boundaries emerge as discrete manifestations of characteristic transitions in continuous information processing dynamics.
\end {abstract}

\begin{keywords}
\noindent
language processing $|$ event boundaries $|$ processing pressure $|$ dynamical systems $|$ large language models
\end{keywords}

\begin{corrauthor}
\texttt{rui.he{@}upf.edu}
\end{corrauthor}

\section*{Introduction}
Everyday experience reaches us as a continuous stream. Scenes change, actions unfold, sounds decay, and interactions move forward without clear physical cuts. Yet cognition does not simply mirror this continuity. It must maintain and integrate incoming information while organizing experience into bounded units. Linguistic experience presents this organization in a grammatically structured form. Words combine into phrases and clauses, and clauses can be recursively embedded within larger constructions. Diverse linguistic traditions have characterized this organization in terms of syntactic constituents and clausal categories, functional and discourse units, and prosodic hierarchies \cite{chomsky2002syntactic, halliday_hallidays_2013, nespor_prosodic_2007}. This organization is also reflected in language processing: behavioral studies show sensitivity to recurring multiword sequences \cite{arnon2010more}, while neural responses track linguistic structures at phrase and sentence scales \cite{pallier_cortical_2011, brennan_abstract_2016, ding_cortical_2016, nelson_neurophysiological_2017, pylkkanen_neural_2019}. Although this process can, in principle, continue without a fixed grammatical limit, actual linguistic constructions eventually come to an end \cite{arsenijevic_absence_2012}. While working memory constraints can limit sentence length \cite{kellogg_working_2004}, they do not by themselves explain why closure occurs at a particular point \cite{christiansen_now-or-never_2016}. One possibility is that linguistic boundaries emerge, at least in part, from the interaction of processing demands as language unfolds over time.

The now-or-never bottleneck view provides a basis for this possibility: transient linguistic input must be rapidly processed and recoded before it is lost \cite{christiansen_now-or-never_2016}. Sentence processing involves demands at both structural and semantic levels \cite{rathi-2021-dependency}. At the structural level, dependency locality theory (DLT) identifies the memory costs of maintaining unresolved dependencies and the integration costs incurred when these relations are resolved \cite{gibson_linguistic_1998, gibson2000dependency}. For example, in 'John, who wears a suit, talked to Mary', 'John' must remain active across the intervening clause until it can be integrated with the predicate 'talked'. Meanwhile, incoming information must be integrated with the evolving semantic context, with processing difficulty depending on contextual expectations \cite{hale_probabilistic_2001, kutas_thirty_2011}: following 'talked to', 'Mary' is more contextually compatible than 'the table'. These demands evolve as information unfolds: unresolved relations must be maintained and eventually integrated, while contextual representations are continually updated. Crucially, structural completion makes closure possible, but does not necessarily determine when it occurs. We propose that the combined effects of these demands accumulate as processing pressure, increasing the likelihood of closure at structurally permissible points.

Under this account, boundaries represent discrete transitions within continuously evolving processing dynamics. Such transitions should involve coordinated changes in processing demands and local representations, rather than a uniform increase in processing difficulty. Existing findings provide clues to these changes. Event boundaries are associated with increased processing time \cite{speer_temporal_2005, zacks_segmentation_2009, smith_differential_2020}, clause and sentence endings produce wrap-up effects \cite{warren_investigating_2009, meister_analyzing_2022}, and boundary words evoke event-related potential signatures associated with memory formation \cite{zogaj_how_2026}. Computational modeling of fMRI responses during narrative comprehension further suggests that information updating between word and sentence levels is concentrated at sentence boundaries \cite{zhou_hierarchical_2025}. Large language models (LLMs) also exhibit boundary-sensitive processing, although such effects are not always strongly expressed in next-token prediction \cite{zou_constituent-constrained_2026}. Changes in hidden-state dynamics, particularly in deeper layers, have instead been linked to sentence-level organization \cite{skrill_large_2023}. These findings suggest that boundaries involve changes in ongoing processing and internal representations, but how such changes relate to the demands that shape boundary formation remains unclear.

We investigate both the formation of boundaries and the processing transitions associated with their occurrence. First, an abstract generative simulation tests whether interacting maintenance, integration, and contextual demands can produce bounded segments without predefined boundary positions or constituent labels. The model generates sentence-like length distributions and distinct interacting timescales, while pressure-dependent closure produces less variable segment lengths than a control simulation where structurally complete segments terminate at a fixed probability. We next test whether corresponding demands predict human boundary placement in continuous speech. In a speech-chunking task, listeners' choices were associated with reduced remaining memory load, elevated preceding memory load, higher integration cost, and greater semantic mismatch, beyond syntactic and prosodic cues. Finally, we examine established sentence boundaries using naturalistic self-paced reading and LLM processing of the same texts. Sentence-final reading time increased alongside changes in processing costs, while the LLM showed pronounced reorganization of internal representational geometry. Explicit boundary status explained additional variation in reading time and hidden-state geometry beyond graded processing demands, with a much smaller contribution to next-token surprisal. Together, these findings support a view of linguistic boundaries as discrete transitions shaped by continuous processing demands and accompanied by changes in representational organization.

\section*{Results}

\subsection*{Emergent boundary from pressure-constrained organization}

\begin{figure*}[h]
\centering
\includegraphics[width=\textwidth]{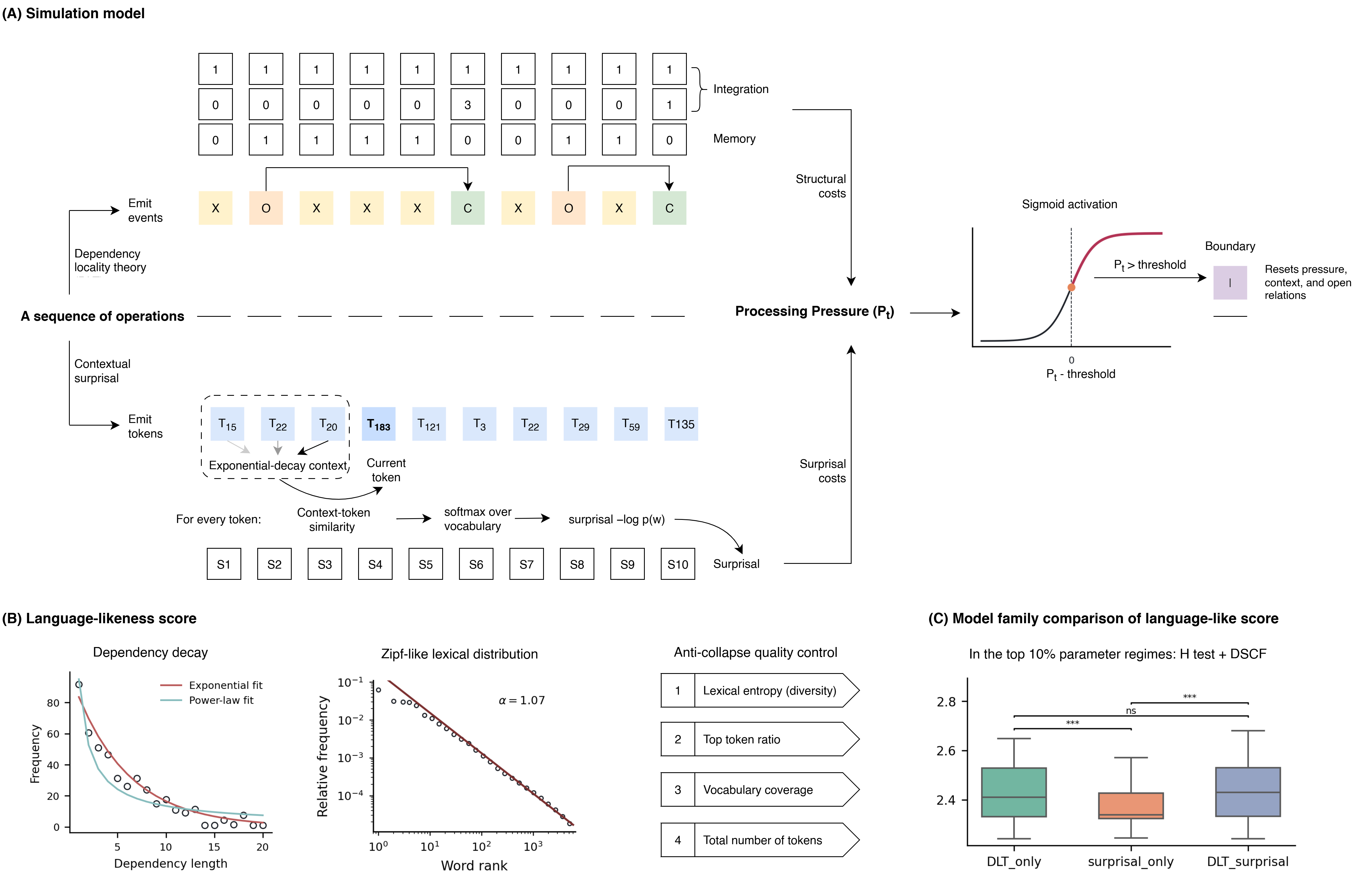}
\caption{
Pressure-constrained simulation and parameter selection.
(A) Simulation model. Dependency-locality costs and contextual surprisal jointly contribute to processing pressure. Accumulated pressure is transformed by a sigmoid function into a probabilistic boundary drive. When open relations remain unresolved, pressure promotes the resolution of open relations. When no relation remains open, pressure increases the probability of a boundary, which resets accumulated pressure and the local contextual states.
(B) Language-likeness score. Parameter regimes were evaluated using dependency-length decay, Zipf-like lexical rank-frequency structure, and anti-collapse lexical quality control.
(C) Top 10\% language-like regimes. Model families were compared within the subset of parameter combinations retained by the language-likeness score.
}
\label{fig:scheme}
\end{figure*}

We first asked whether boundary-like organization could arise from local processing demands without predefined sentence boundaries or constituent labels. To test this, we developed an abstract generative model with interacting structural and contextual streams (Figure~\ref{fig:scheme}A). The structural stream generated operations that opened, maintained, and resolved dependency-like relations. Drawing on dependency locality theory (DLT), we distinguished memory cost, reflecting the number of unresolved relations held active, from integration cost, incurred when these relations were resolved. In DLT, integration cost reflects both the distance between dependent elements and the introduction of new discourse referents. Because our synthetic model contained neither lexical categories nor explicit discourse referents, we treated each generated token as a minimal discourse unit and approximated integration cost by the number of intervening units crossed when a relation was resolved. This captures the distance-sensitive component of DLT without implementing the full theory. In parallel, the contextual stream operated over a synthetic semantic space in which each item in a finite pseudo-vocabulary was assigned a randomly initialized vector sampled around latent semantic centers, allowing nearby items to share semantic similarity. Preceding token representations were combined into an exponentially decaying context state that assigned greater weight to recent input. Tokens were generated according to their compatibility with this evolving state, and contextual surprisal quantified how unexpected each new token was given the preceding context.

At each step, memory cost, integration cost, and contextual surprisal were combined with residual pressure carried over from earlier steps. The resulting processing pressure was transformed through a sigmoid function into a probabilistic boundary drive. When relations remained open, increasing pressure promoted their closure. Once all relations had been resolved, the same accumulated pressure increased the probability of emitting a boundary. Boundary emission reset pressure, contextual state, and open relations, allowing a new segment to begin. Thus, segment boundaries were determined by the interaction of processing demands and structural operations rather than supplied as predefined positions.

We explored a broad parameter space, evaluating each configuration across ten random seeds (SI, Tables S1 and S2). To identify non-degenerate regimes without requiring the model to reproduce natural language in detail, we constructed a language-likeness score based on dependency-length decay \cite{chen-gerdes-2019-relation}, approximately Zipf-like lexical rank-frequency distributions \cite{piantadosi2014zipf}, and controls against lexical collapse (Figure~\ref{fig:scheme}B).

The parameter scan identified regimes that reproduced basic statistical regularities of natural language without predefined linguistic boundaries. We compared three model families differing in the sources of processing pressure: memory and integration costs alone (DLT-only), contextual surprisal alone (surprisal-only), or all three combined (DLT--surprisal). Among the top 10\% of configurations ranked by language-likeness, the combined model achieved the highest numerical overall scores (Figure~\ref{fig:scheme}C). Configurations incorporating memory and integration costs scored higher than those driven by contextual surprisal alone. Compared with DLT-only configurations, adding contextual surprisal did not significantly improve the overall language-likeness score within this selected subset, although it was associated with a closer fit to Zipf-like lexical distributions (SI, Figure S1). High-scoring configurations were found across a range of boundary thresholds, pressure carry-over values, contextual strengths, and structural-cost weights (SI, Figure S2). These results show that pressure-dependent segmentation can generate basic language-like statistical properties across multiple parameter settings. We selected the highest-scoring combined configuration (SI, Table S3) for subsequent analysis of its boundary dynamics.

\subsection*{Pressure-release transitions}
Having identified the highest-scoring combined parameter regime, we next examined how processing demands evolved around the boundaries generated by the selected model. The configuration was rerun across 500 random seeds, and model variables were aligned to emitted boundaries. Position 0 corresponded to the final pre-boundary event, and position +1 to the first post-boundary event. For comparison, an equal number of non-boundary control positions were sampled from the same sequences, at least three positions away from any boundary.

\begin{figure}[ht]
\centering
\includegraphics[width=\linewidth]{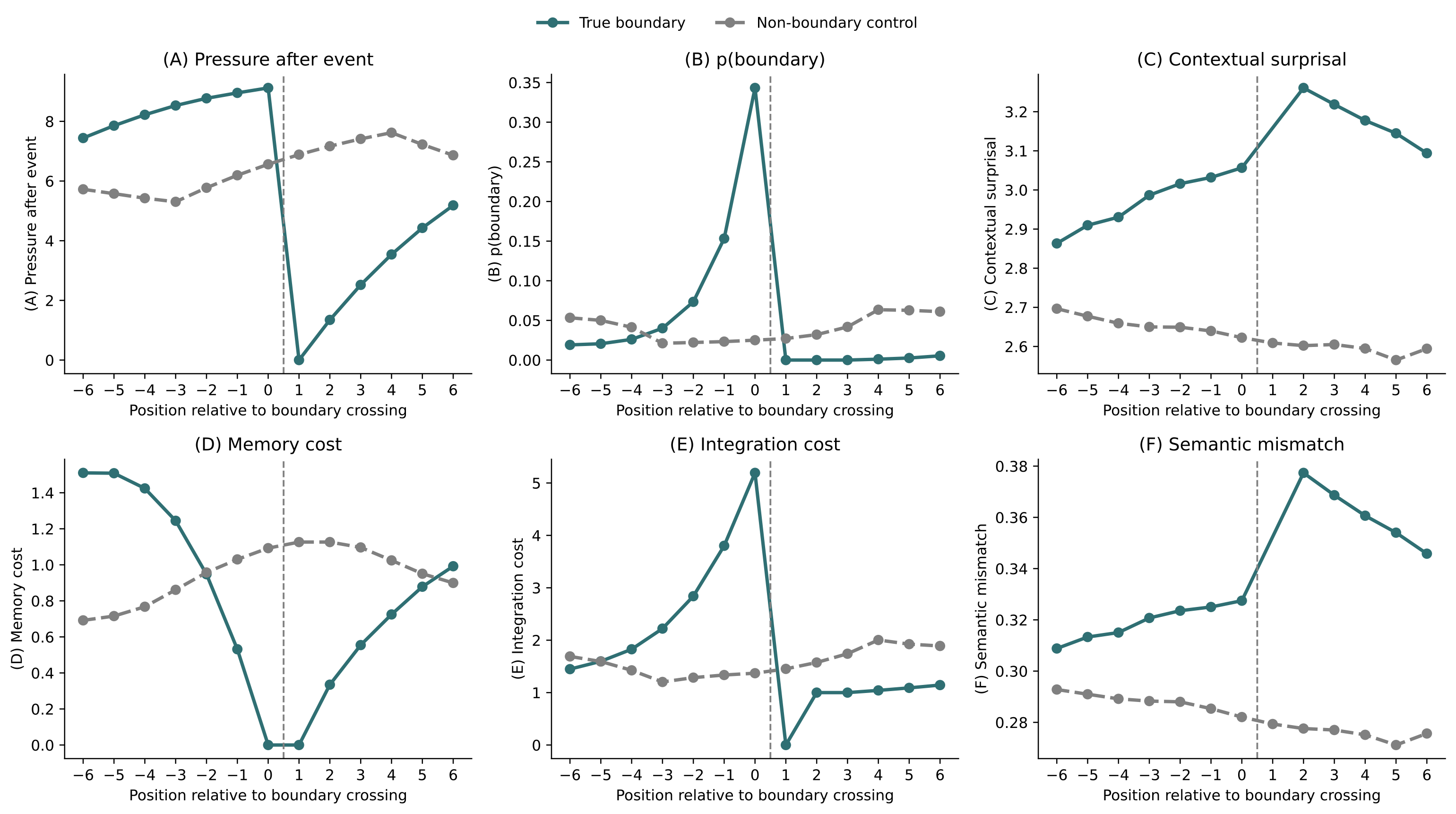}
\caption{
Boundary-aligned profiles in the best-performing pressure-constrained model.
Analyses were performed on the highest-scoring DLT-surprisal parameter regime, rerun across 500 random seeds. Model variables were aligned to emitted boundaries and compared with non-boundary control positions sampled from the same simulations. The vertical dashed line marks the boundary crossing. Six variables of interest: (A) Processing pressure; (B) Boundary probability; (C) Contextual surprisal costs; (D) Memory costs; (E) Integration costs; and (F) Semantic mismatch. 
}
\label{fig:boundary_align}
\end{figure}

Consistent with the model's boundary-generation rule, accumulated processing pressure and boundary probability increased toward boundary emission and dropped sharply afterward (Figure~\ref{fig:boundary_align}A,B). These transitions were absent around non-boundary control positions. Beyond this expected pressure reset, we examined whether the individual processing demands exhibited coordinated changes around boundaries.

The structural and contextual components showed distinct temporal profiles. As boundaries approached, open relations were progressively resolved, resulting in declining memory cost and increasing integration cost, which peaked around boundary emission (Figure~\ref{fig:boundary_align}D,E). Contextual surprisal and semantic mismatch exhibited more gradual changes, increasing around boundaries and peaking shortly after emission (Figure~\ref{fig:boundary_align}C,F). These contextual changes were broader and delayed relative to the sharper structural closure and integration profiles. No comparable coordination was observed around non-boundary control positions.

Although structural completion and pressure reset were specified in the model, the detailed temporal profiles of processing demands were not. The simulation produced distinct structural and contextual dynamics around closure, with sharper changes in dependency-related costs and more gradual contextual changes. These coordinated patterns emerged from the interaction of local processing operations rather than from predefined temporal profiles.

\begin{figure}[htbp]
\centering
\includegraphics[width=\linewidth]{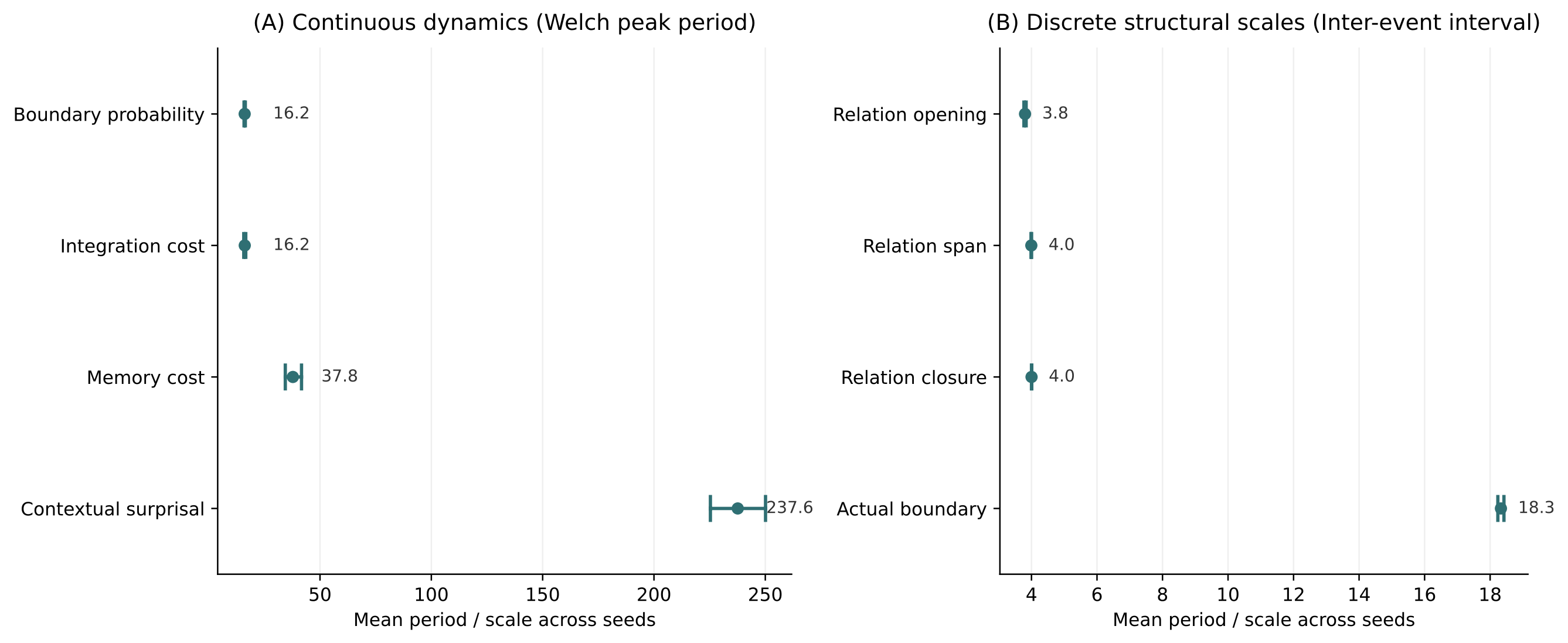}
\caption{
Temporal scales of continuous dynamics and discrete structural events in the selected model.
Analyses were performed on the highest-scoring DLT--surprisal parameter regime rerun across 500 random seeds. Points show mean period or scale across seeds. Error bars indicate 95\% bootstrap confidence intervals.
(A) Dominant periods of continuous model variables estimated from Welch peak periods. Boundary probability and integration cost showed similar segment-scale periods, whereas memory cost varied over a longer scale and contextual surprisal over a much slower scale.
(B) Discrete structural scales estimated from median inter-event intervals or relation spans. Relation openings, relation closures, and relation spans occurred at short local scales, whereas actual boundaries occurred at a longer segmental scale.
}
\label{fig:timescales}
\end{figure}

\subsection*{Distinct characteristic timescales in model dynamics}

Beyond these boundary-centered changes, we examined whether different processing demands operated over distinct characteristic timescales (Figure~\ref{fig:timescales}). For continuous variables, timescales were estimated from the dominant non-zero frequencies in Welch power spectra. For discrete structural events, they were estimated from median inter-event intervals or relation spans.

The continuous variables exhibited distinct temporal profiles (Figure~\ref{fig:timescales}A). Boundary probability and integration cost showed similar dominant timescales of approximately 16 events, suggesting that boundary drive and the integration of resolved relations varied at a common segmental scale. Memory cost varied over a longer timescale of approximately 38 events, reflecting slower fluctuations in the load of unresolved relations. Contextual surprisal was dominated by still lower frequencies, indicating that contextual compatibility varied over a broader temporal window than the structural processing demands.

Discrete structural events revealed a complementary organization (Figure~\ref{fig:timescales}B). Relation openings, closures, and spans occurred at short local scales of approximately four events, whereas emitted boundaries were separated by approximately 18 events. Inter-boundary intervals followed a strongly right-skewed, long-tailed distribution (SI, Figure S3), qualitatively resembling natural sentence-length distributions \cite{borbely-kornai-2019-sentence}.

Together, these findings reveal a multiscale organization in which local dependency operations occur at short timescales, integration and boundary formation vary at an intermediate segmental scale, memory load fluctuates over longer periods, and contextual information evolves most slowly. Thus, the pressure-constrained model generates distinct but interacting temporal scales of linguistic-like organization without imposing an explicit hierarchy of linguistic units.

\subsection*{Processing pressure beyond structural completion}

To assess the contribution of pressure-dependent boundary formation, we compared the model with a completion-only control in which boundaries occurred with a fixed, pressure-independent probability once all open relations had been resolved. The two models were calibrated to produce comparable mean segment lengths. Pressure-dependent closure yielded a more constrained segment-length distribution and a sharper increase in integration cost preceding boundaries, whereas completion-only closure produced greater length variability and weaker boundary-centered coordination of processing demands (SI, Figure S4). These findings suggest that processing pressure contributes to determining when structurally permissible boundaries are realized.

\subsection*{Processing costs predict human boundary choices}
The simulation revealed coordinated changes in structural and contextual processing demands around generated boundaries. To test whether corresponding structural and contextual demands predict human boundary choices in natural speech, we analyzed an independent perceptual chunking dataset \cite{vetchinnikova_chunking_2023}. In the original experiment, 50 listeners marked perceived boundaries while listening to extracts of spontaneous speech. The accompanying transcripts were presented without sentence punctuation, so listeners marked perceived boundaries without written punctuation cues. Across 5,140 candidate word positions, we quantified memory cost remaining after each word, integration cost associated with dependency resolution, and local semantic mismatch. 

Structural costs were derived from dependency parses produced by the Stanza English parser \cite{qi_stanza_2020}. Following the distinction in DLT \cite{gibson2000dependency}, memory cost indexed the number of dependency relations remaining unresolved after each word. Integration cost summed the distance costs of relations resolved at the current word, measured by the number of intervening discourse referents. Nouns, proper nouns, and verbs were treated as content-bearing referential units.

Lexical-semantic mismatch was estimated separately using static fastText embeddings \cite{grave_learning_2018}. We used static word embeddings to capture distributional semantic relationships between words, without introducing the sentence-specific syntactic and contextual information encoded by contextualized embeddings. For each word, semantic mismatch was defined as the cosine dissimilarity between its embedding and the mean embedding of the preceding four words. This provided a local measure of lexical-semantic dissimilarity independent of the dependency-derived structural costs.

\begin{figure}[htbp]
\centering
\includegraphics[width=\linewidth]{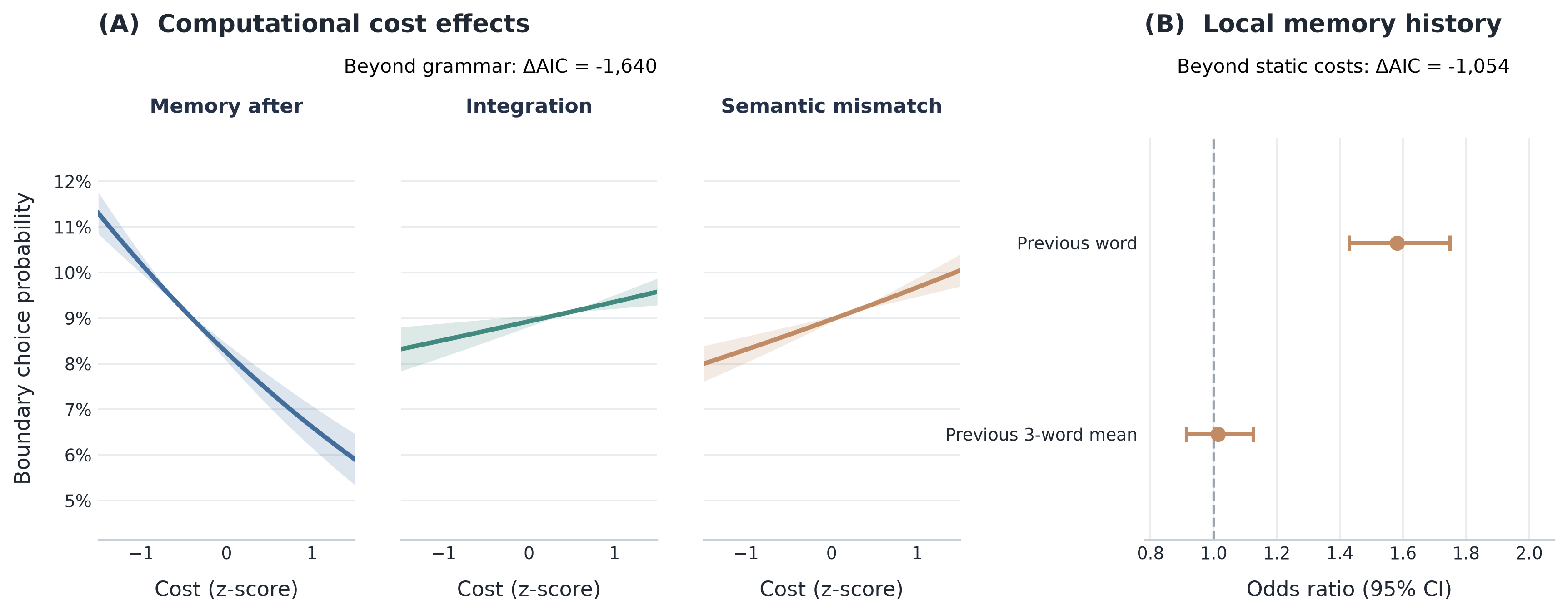}
\caption{Processing costs and local memory history in human boundary judgments. (A) Adjusted boundary-choice probabilities across standardized memory cost after the current word, integration cost, and local semantic mismatch. Shading indicates 95\% confidence intervals. (B) Odds ratios for memory cost at the preceding word and mean memory cost across the preceding three words. Error bars indicate 95\% confidence intervals; the dashed line marks an odds ratio of 1.}
\label{fig:human_chunking}
\end{figure}

Adding these predictors substantially improved model fit (Akaike information criterion, $\Delta$AIC = $-1,639.63$; joint Wald $p<0.001$; Figure~\ref{fig:human_chunking}A). All three costs independently predicted boundary choices. Higher post-word memory cost was associated with lower odds of a boundary choice (odds ratio [OR] = 0.689, $p<0.001$), while higher integration cost (OR = 1.094, $p=0.002$) and semantic mismatch (OR = 1.158, $p<0.001$) were associated with increased odds. To ensure that these effects were not driven by the endpoints of the largest syntactic units identified in the original annotations, where memory cost necessarily reached zero, we repeated the analysis after excluding these positions. All three associations remained significant (SI, Figure S5), and the results were also robust to treating grammatical boundary strength as a categorical rather than a continuous predictor (SI, Figure S6).

We further examined whether boundary choices depended on recent changes in memory load rather than its current level alone. Adding memory cost at the preceding word and the mean cost across the preceding three words further improved model fit ($\Delta$AIC = $-1,054.40$, joint Wald $p<0.001$; Figure~\ref{fig:human_chunking}B). Memory cost at the immediately preceding word positively predicted boundary choices after accounting for current processing demands (OR = 1.581, $p<0.001$), while the three-word mean showed no additional association (OR = 1.014, $p=0.801$).

These findings provide behavioral support for the predicted relationships between structural and contextual processing demands and boundary placement, broadly consistent with the simulation.

\subsection*{Human reading shows a pressure-release profile at sentence boundaries}

The speech-chunking task recorded listeners' boundary choices but did not measure reading time, which provides a complementary behavioral measure of how processing difficulty evolves as a sentence unfolds. We therefore analyzed word-by-word self-paced reading time from the Naturalistic Stories corpus \cite{futrell_natural_2021} to examine processing dynamics around established sentence boundaries. To reduce spillover from preceding words, we residualized log-transformed reading time against previous-word log reading time and story identity \cite{vasishth2006proper}.

\begin{figure*}[htbp]
\centering
\includegraphics[width=\linewidth]{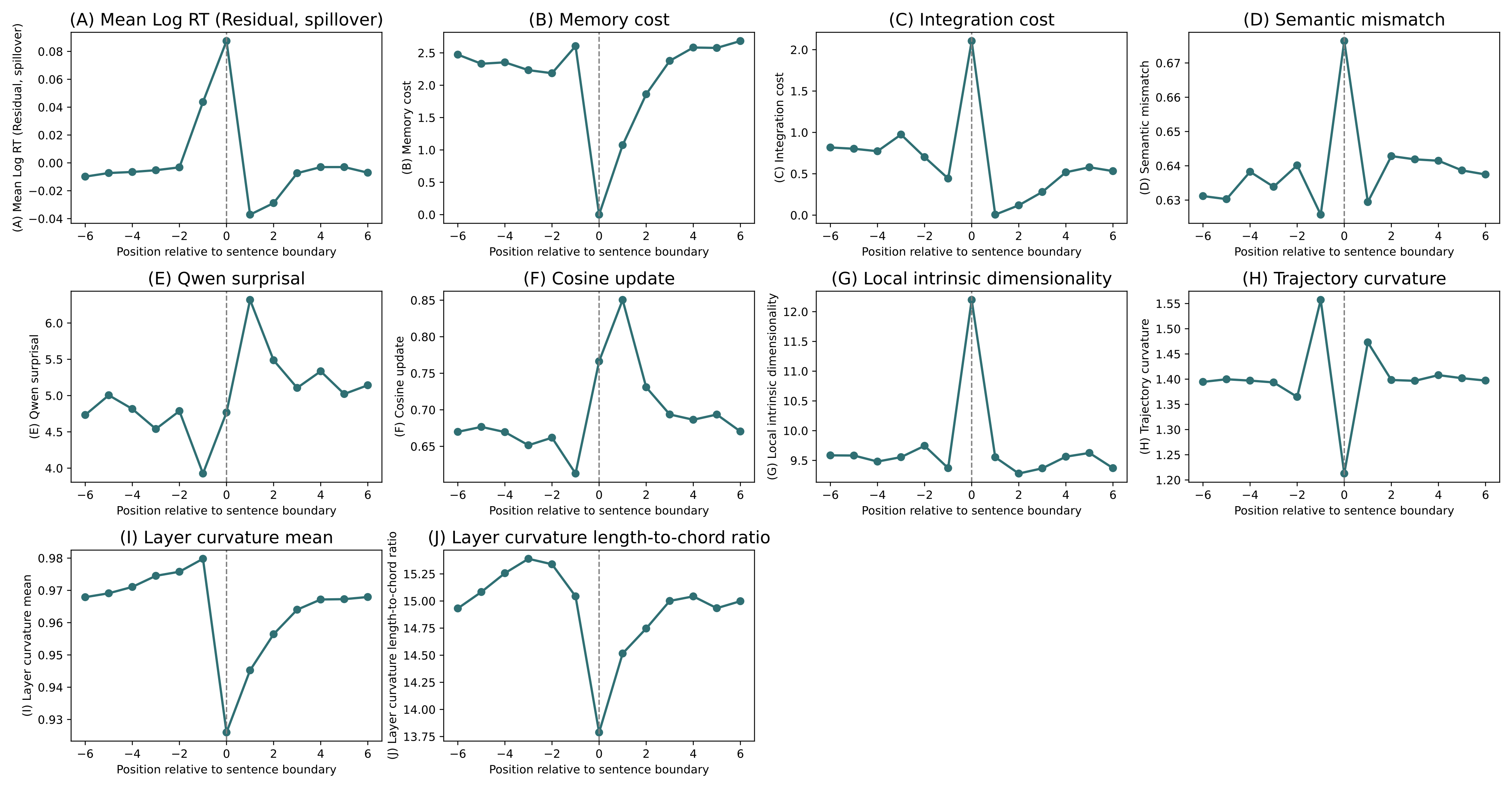}
\caption{
Sentence-boundary-aligned profiles of human reading time, model-derived surprisal, and LLM measures.
All measures were aligned to sentence boundaries in the naturalistic stories. The vertical dashed line marks the sentence boundary position.
(A) Mean log reading-time residuals, including spillover correction.
(B) Memory cost from dependency parsing.
(C) Integration cost from dependency parsing.
(D) Semantic mismatch increased at the boundary.
(E) LLM surprisal.
(F) Cosine distance between consecutive hidden states.
(G) Local intrinsic dimensionality.
(H) Trajectory curvature among tokens.
(I) Averaged layer curvature.
(J) Length-to-chord ratio of layer curvature.
}
\label{fig:rt_llm_boundary_profiles}
\end{figure*}

Residual reading time showed a clear sentence-boundary profile. It increased as the boundary approached, peaked at the sentence-final word, and dropped immediately afterward (Figure~\ref{fig:rt_llm_boundary_profiles}A). This pattern is consistent with a sentence-final wrap-up effect, in which readers allocate additional processing time at clause or sentence boundaries to integrate or close the preceding material \cite{warren_investigating_2009}.

To examine the processing demands accompanying this behavioral profile, we estimated dependency-based memory and integration costs and local semantic mismatch using the same procedures as in the chunking analysis. Memory cost declined toward sentence endings, as expected from dependency parsing within sentences. Integration cost peaked at the sentence-final word and dropped immediately afterward (Figure~\ref{fig:rt_llm_boundary_profiles}B,C). Semantic mismatch also reached its maximum at the sentence boundary (Figure~\ref{fig:rt_llm_boundary_profiles}D).

Together, these profiles indicate that the sentence-final increase in reading time coincided with the resolution of open dependencies, elevated integration demands, and increased local semantic mismatch. The structural dynamics closely resembled those observed in the simulation, with declining memory load and integration costs concentrated around closure. A temporal difference emerged in the contextual component: semantic mismatch peaked at the sentence-final word in naturalistic reading, rather than immediately after boundary emission as in the simulation. Thus, the human reading data revealed a boundary-centered processing profile that broadly paralleled the simulated pressure-release dynamics, while differing in the timing of contextual changes.

\subsection*{LLM representations reorganize at sentence boundaries}
We also examined how sentence boundaries affect prediction and internal representations in a large language model. Using Qwen3-4B \cite{qwen3technicalreport}, we processed the same naturalistic stories and measured next-token surprisal, word-to-word representational updates, and hidden-state geometry across words and model layers.

At the output level, contextual surprisal increased toward sentence boundaries and peaked immediately afterward (Figure~\ref{fig:rt_llm_boundary_profiles}E), resembling the delayed contextual changes observed in the simulation (Figure~\ref{fig:boundary_align}C,F). This post-boundary increase in prediction difficulty suggested a transition in contextual processing, which we further examined through the model's hidden-state geometry.

Cosine distance between consecutive token representations, measuring the magnitude of word-to-word representational updates, also peaked immediately after the boundary (Figure~\ref{fig:rt_llm_boundary_profiles}F). In contrast, local intrinsic dimensionality (ID), reflecting the effective dimensionality of the local representational neighborhood \cite{tsukagoshi_redundancy_2025}, reached its maximum at the sentence-final word (Figure~\ref{fig:rt_llm_boundary_profiles}G). These profiles suggest that local representational complexity was greatest at closure, followed by a pronounced update as the model entered the next sentence.

We further examined the geometry of representational trajectories. Temporal trajectory curvature, which captures changes in the direction of successive token representations, increased immediately before the boundary, dropped to a minimum at the sentence-final word, and recovered afterward (Figure~\ref{fig:rt_llm_boundary_profiles}H). The boundary was therefore associated with a transient straightening of the representational trajectory following increased pre-boundary turning.

A similar pattern emerged across model depth. Mean layer curvature and the layer-wise length-to-chord ratio, characterizing the geometry of each token's transformation across successive layers, both decreased at sentence-final positions and recovered afterward (Figure~\ref{fig:rt_llm_boundary_profiles}I,J). These decreases indicate that layer-wise transformations became more direct at closure, paralleling the reduction in temporal trajectory curvature. Sentence boundaries were thus associated with transient straightening along both the temporal and layer-wise dimensions of representational change.

Together, these measures revealed distinct but coordinated boundary-centered dynamics. Local intrinsic dimensionality and representational straightening were concentrated at the sentence-final word, while surprisal and the magnitude of token-to-token updates peaked immediately afterward. Rather than producing a uniform change in representational activity, sentence boundaries were associated with a temporally structured transition between the processing of the preceding sentence and the representation of the next.

\begin{figure*}[t!]
\centering
\includegraphics[width=\linewidth]{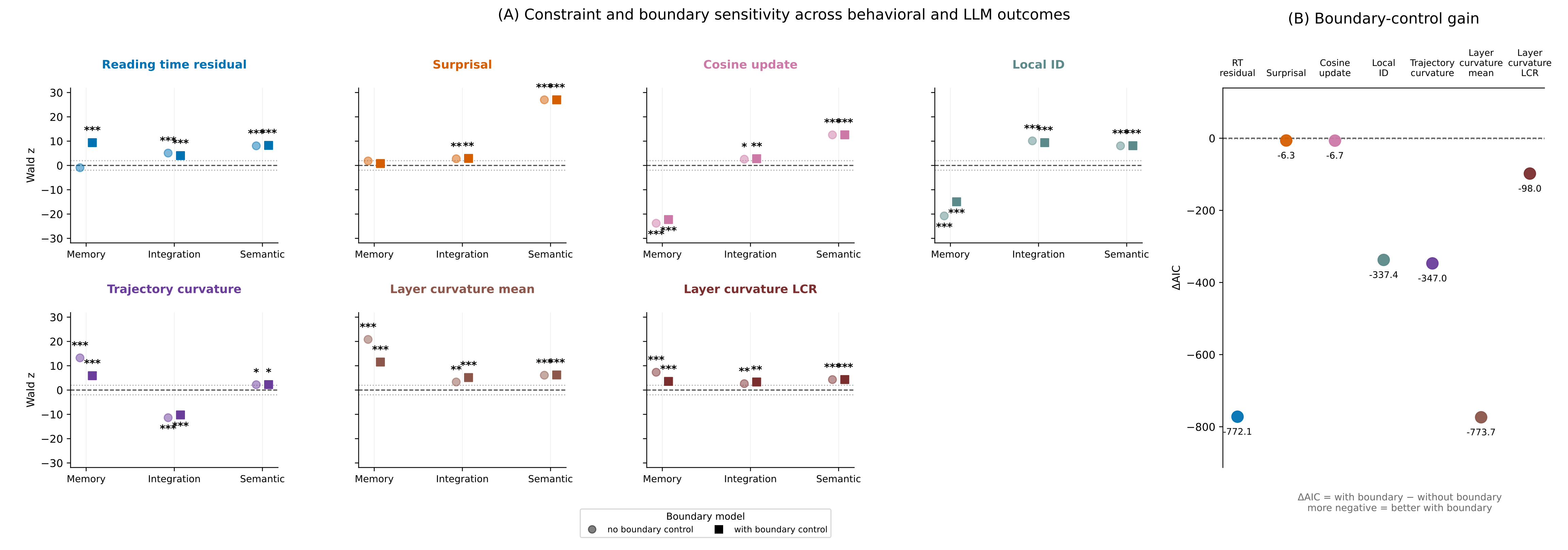}
\caption{
(A) Constraint sensitivity across behavioral and LLM outcomes. Each panel shows the Wald $z$-statistics for memory cost, integration cost, and static semantic mismatch as predictors of one outcome. Outcomes included spillover-adjusted reading-time residuals, Qwen surprisal, cosine update, local intrinsic dimensionality, trajectory curvature, mean layer curvature, and layer length-to-chord ratio (LCR). Circles show models without explicit sentence-boundary control. Squares show models including sentence-boundary status as an additional categorical predictor. Stars indicate statistical significance corrected by False Discovery Rate.
(B) Boundary-control gain. Points show $\Delta\mathrm{AIC}$, defined as AIC with boundary control minus AIC without boundary control. More negative values indicate better model fit after adding explicit sentence-boundary status. 
}
\label{fig:regression}
\end{figure*}

\subsection*{Graded pressure and discrete closure make distinct contributions across outcomes}

The boundary-aligned profiles revealed distinct changes in human reading time and LLM processing at sentence endings. We tested whether these changes could be explained by graded processing demands or whether categorical sentence-boundary status accounted for additional variation. To distinguish these contributions, we compared two regression models for each outcome. The first included memory cost, integration cost, and semantic mismatch as graded predictors that varied across words. The second additionally included categorical boundary status, indicating whether a word was sentence-final. Both models controlled for word position and story identity (Figure~\ref{fig:regression}). Adding boundary status improved model fit across all outcomes, although the magnitude of this improvement varied substantially (Figure~\ref{fig:regression}B; SI, Figure S7).

Human reading time was positively associated with integration cost and semantic mismatch, with memory cost also showing a positive association after boundary status was included (Figure~\ref{fig:regression}A). Boundary status produced a substantial reduction in AIC, indicating that the selected continuous processing costs did not fully account for sentence-final reading difficulty. This additional association need not imply an independent boundary-generating mechanism, as categorical boundary status may also capture sentence-final properties not represented by the continuous predictors.

LLM surprisal showed a different pattern. Semantic mismatch was its strongest predictor, while memory and integration costs showed comparatively weaker associations (Figure~\ref{fig:regression}A). Adding boundary status yielded only a small improvement in model fit (Figure~\ref{fig:regression}B), suggesting that output-level prediction difficulty was primarily associated with graded lexical-semantic demands.

The LLM's internal representations showed greater sensitivity to boundary status. Although cosine update was associated with all three processing costs, its additional improvement from boundary control remained small. In contrast, local intrinsic dimensionality, trajectory curvature, mean layer curvature, and layer length-to-chord ratio showed substantially larger reductions in AIC after boundary status was included (Figure~\ref{fig:regression}B). Thus, several aspects of hidden-state geometry captured boundary-related variation beyond that explained by the measured processing costs.

This dissociation distinguishes the graded demands preceding closure from the representational changes associated with the boundary itself. Human reading difficulty and LLM representational geometry retained substantial boundary-related effects, while LLM surprisal predominantly tracked lexical-semantic mismatch. The results suggest that sentence closure is expressed differently in observable processing difficulty: next-token prediction and internal representational dynamics.

\section*{Discussion}
This study examined how linguistic boundaries emerge from interacting processing demands and how closure is expressed during language processing. Our generative simulation showed that structural completion made closure possible, while accumulating memory, integration, and contextual demands shaped when it occurred. The model produced bounded segments with sentence-like length distributions and coordinated changes in processing demands around boundaries. We examined two implications of these findings in natural language. First, corresponding processing demands predicted listeners' boundary choices in spontaneous speech beyond grammatical and prosodic cues, providing behavioral support for the proposed account of boundary formation. Second, the simulation predicted characteristic changes in processing demands around closure. In human self-paced reading, sentence-final slowing coincided with changes in memory load, integration cost, and semantic mismatch, broadly consistent with the simulated profiles. Analyzing the same texts with an LLM allowed us to examine the associated internal representations, which showed pronounced geometric changes at sentence boundaries. Moreover, categorical boundary status explained additional variation in human reading time and LLM representational geometry beyond graded processing demands, but contributed much less to next-token surprisal. Together, these findings suggest that linguistic boundaries can emerge from interacting processing demands and that their occurrence marks a characteristic transition in ongoing processing.

This account complements formal descriptions of linguistic structure. Constituency and dependency representations characterize how linguistic elements are organized and the structural conditions under which a sentence is complete. A distinct mechanistic question is why an unfolding sequence closes at one structurally permissible point rather than continuing into a larger structure. Our model addresses this question through the interaction of dependency resolution and processing pressure: the former makes closure possible, while accumulating structural and contextual demands regulate when it occurs. The completion-only control reinforces this distinction, showing that structural completion permits segmentation, but pressure-dependent gating further shapes segment-length distributions and boundary-centered dynamics. Human boundary choices provide converging evidence, with processing costs predicting boundary placement beyond grammatical and prosodic cues. Sentence closure thus reflects the interaction of grammatical constraints and processing dynamics, rather than either alone. Although our analysis focuses on sentence-level closure, the distinct characteristic timescales observed in the simulation suggest that related organizing principles may extend to other levels of linguistic organization.

Linguistic closure in the simulation was organized across multiple characteristic timescales. Local dependency operations occurred over approximately four generated units, while integration cost and boundary drive varied over a segmental scale of 16–18 units. Memory load evolved over longer intervals, and contextual surprisal showed the slowest variation. The resulting segment lengths exhibited a right-skewed, long-tailed distribution, qualitatively resembling natural sentence-length distributions \cite{borbely-kornai-2019-sentence}. This multiscale organization suggests that larger linguistic units can arise through the coordination of local structural operations and more persistent contextual demands, without requiring a fixed chunking window or a predefined hierarchy of units \cite{ding_sequence_2025}. Such coordination is consistent with local capacity constraints on online processing \cite{n_magical_2001, lo_periodic_2023} and the accumulation of information across multiple temporal windows \cite{hasson_hierarchy_2008, honey_slow_2012, clewett_transcending_2019, staub_coherence_2026}. It also suggests a possible connection to metastability. A boundary occurred when the current configuration was brought to completion and the system transitioned to a new one. In this limited sense, the results are compatible with a metastable view of language processing: sentence-level configurations remain temporarily organized without becoming permanently fixed and are reorganized when the demands sustaining the current state reach a transition point and move the system beyond its current stable range. We suggest this interpretation cautiously and as provisional, as formal metastability was not directly quantified.

The role of processing dynamics in boundary formation raises a further question about what happens once closure is realized. In the speech-chunking analysis, processing costs remained predictive after excluding top-level unit endings, whereas sentence-final memory release in the reading analysis was partly constrained by sentence-level parsing. Nevertheless, sentence-final slowing coincided with elevated integration cost and semantic mismatch. Moreover, categorical boundary status improved model fit beyond these graded processing demands, suggesting that the measured costs did not fully capture the processing changes associated with closure.

LLM next-token surprisal showed a different pattern. Although surprisal peaked immediately after sentence boundaries, it was most strongly associated with lexical-semantic mismatch, and adding boundary status yielded only a small improvement in model fit. Surprisal quantifies the unexpectedness of incoming words under the model's current context, but may not fully capture the computations associated with closure itself. This interpretation is consistent with evidence that next-token prediction can be comparatively insensitive to segment boundaries \cite{zou_constituent-constrained_2026}.

The internal representations of the LLM revealed a more pronounced signature of sentence closure. Boundary status accounted for substantial additional variation in local intrinsic dimensionality and representational curvature, even though its contribution to output surprisal was relatively small. These geometric measures also showed a distinct temporal organization: local intrinsic dimensionality peaked at sentence-final positions, representational trajectories became straighter around closure, and token-to-token updates reached their maxima immediately afterward. Sentence boundaries thus coincided with a reorganization of internal representations that was not fully captured by next-token prediction. This provides a possible computational counterpart to the increased processing demands observed at sentence endings in human reading.

The reduction in layer-wise curvature further raises the possibility that representational closure reflects the progressive satisfaction of grammatical constraints. Previous studies have linked curvature across transformer layers to contextual processing \cite{sipio_curved_2025, king_representational_2026}, while grammatical structure has been shown to constrain the organization of contextual meaning \cite{he_grip_2026}. As a sentence unfolds, dependency resolution and the narrowing of possible interpretations may progressively constrain its internal representation, leading to more direct transformations across model layers at closure. The observed straightening may therefore reflect a convergence of structural and contextual constraints as the sentence reaches a locally complete configuration. This interpretation connects representational geometry to the interaction between grammatical organization and incremental language processing \cite{gibson_linguistic_1998, levy_expectation-based_2008}, suggesting a potential mechanism through which grammatical constraints shape the dynamics of sentence closure.

Several limitations qualify these findings. The simulation deliberately simplified linguistic structure and meaning through synthetic semantic representations, abstract dependency operations, and an approximate integration-cost measure. Pressure-dependent boundary gating and post-boundary resets were specified by design, while the resulting segment distributions and component-specific dynamics arose from their interaction with local processing operations. The completion-only comparison tested a specific alternative based on fixed termination probability, rather than all possible mechanisms of structural closure. The human analyses were observational: processing costs predicted boundary choices beyond grammatical and prosodic cues, but these associations do not establish causality. Similarly, sentence-boundary effects in reading time and LLM measures may reflect additional linguistic properties not captured by our predictors. Dependency-derived costs and semantic mismatch were selective approximations of processing demands, and the LLM findings were limited to written English narratives and a single model. Extending these analyses across languages, model architectures, and linguistic levels will help establish the generality of the proposed mechanism.

In conclusion, the present findings support a mechanistic account of linguistic closure in which structural completion makes boundaries possible, while interacting memory, integration, and contextual demands contribute to when they are realized. These dynamics organize segmentation across multiple timescales, predict human boundary choices, and are reflected in distinct behavioral and representational changes around sentence boundaries. Sentence closure thus emerges as a transition in ongoing language processing, linking the organization of linguistic structure to the dynamics through which it is realized.

\section*{Materials and methods}

\subsection*{Simulation model}
We implemented an abstract pressure-constrained sequence model to test whether boundary-like organization could emerge from local structural and contextual demands without pre-specified sentence boundaries or constituent labels. The model generated two coupled streams: a structural stream representing the opening, maintenance, and closure of relations, and a lexical stream representing context-sensitive token generation. Full implementation details and parameter definitions are provided in the Supplementary Information.

The structural stream consisted of relation-opening, relation-maintaining, and relation-closing operations. At each step $t$, memory cost ($M_t$) was defined as the number of currently open relations. Integration cost ($I_t$) was computed when a relation was closed. Following the logic of dependency locality theory (DLT), integration cost increased with the number of intervening generated units crossed by the resolved relation. Because the synthetic model did not contain explicit representations of discourse referents, each generated unit was treated as a minimal discourse unit. Integration cost therefore provided a simplified analogue of discourse-referent distance.

The lexical stream was generated from a finite vocabulary embedded in a synthetic semantic space. Before generating token $w_t$, the model maintained an exponentially decaying context state,
\begin{equation*}
m_t = (1-\eta)m_{t-1} + \eta v_{w_{t-1}},
\end{equation*}
where $v_{w_{t-1}}$ is the embedding of the preceding token and $\eta$ controls the context update rate. The probability of each candidate token was computed from its similarity to the current context state,
\begin{equation*}
p(w_t = w \mid m_t) =
\frac{\exp\left(\beta_{\mathrm{ctx}}\,\mathrm{sim}(m_t, v_w)\right)}
{\sum_{u \in V} \exp\left(\beta_{\mathrm{ctx}}\,\mathrm{sim}(m_t,v_w)\right)},
\end{equation*}

where $V$ is the vocabulary, $\mathrm{sim}(\cdot,\cdot)$ denotes cosine similarity, and $\beta_{\mathrm{ctx}}$ controls the strength of contextual constraint. Contextual surprisal was then defined as
\begin{equation*}
S_t = -\log p(w_t \mid m_t).
\end{equation*}

Total processing pressure combined the structural and contextual terms:
\begin{equation*}
P_t = \rho P_{t-1} + a_{\mathrm{mem}} M_t + b_{\mathrm{int}} I_t + \gamma S_t,
\end{equation*}
where $\rho$ controls pressure carry-over, $a_{\mathrm{mem}}$ and $b_{\mathrm{int}}$ weight memory and integration costs, and $\gamma$ weights contextual surprisal. Processing pressure was transformed into a boundary drive using a sigmoid function,
\begin{equation*}
q_t = \sigma\left(k(P_t-\theta)\right),
\end{equation*}
where $\theta$ specifies the midpoint of the sigmoid transformation, $k$ controls its steepness, and $\sigma$ denotes the sigmoid function. When no relations remained open, $q_t$ entered a categorical event distribution as the weight assigned to boundary emission. The corresponding weights for an ordinary event and the opening of a new relation were $1$ and $b_{\mathrm{open}}(1-q_t)$, respectively. These weights were normalized before an event was sampled. When one or more relations remained open, boundary emission was prohibited, and increasing $q_t$ instead increased the weight assigned to relation closure. Boundaries were therefore emitted stochastically as a function of accumulated pressure, rather than deterministically whenever pressure crossed $\theta$. Boundary emission reset accumulated pressure and the local contextual state. Because boundaries were permitted only after all open relations had been resolved, their locations were determined by the evolving interaction among structural load, integration demand, contextual mismatch, and pressure accumulation, rather than supplied as target labels.

\subsubsection*{Language-likeness score}
Each parameter configuration was simulated using ten random seeds. Each resulting synthetic corpus was evaluated using a language-likeness score. The score combined three broad criteria: dependency-length decay \cite{chen-gerdes-2019-relation}, Zipf-like lexical rank-frequency structure \cite{piantadosi2014zipf}, and anti-collapse quality control. 

The dependency component assessed whether the frequency of generated relations decreased with relation length and whether the resulting dependency-length distribution was compatible with the exponential or power-law-like decay observed in natural language \cite{chen-gerdes-2019-relation}. The lexical component assessed whether token rank and frequency followed an approximately Zipf-like relationship \cite{piantadosi2014zipf}.

Quality-control components were used to identify degenerate simulations. These included simulations dominated by a single token, simulations with insufficient lexical diversity, and simulations that produced too little material for reliable estimation. Each quality-control measure was transformed into a bounded score between 0 and 1 according to a pre-specified heuristic threshold. The final language-likeness score was obtained by summing the dependency, lexical, and quality-control component scores. Component scores were first calculated separately for each seed and then averaged across the ten seeds to obtain one score for each parameter configuration. Full definitions and scoring functions are provided in the SI.

\subsubsection*{Model-family comparison}
We compared three model families: DLT-only models, surprisal-only models, and combined DLT–surprisal models. DLT-only models included memory and integration costs but no contextual-surprisal contribution to pressure. Surprisal-only models included contextual surprisal but no memory or integration contribution, whereas combined models included all three sources of processing pressure. Parameter configurations were ranked by their mean language-likeness score, and the top 10\% were retained to characterize the highest-performing region of parameter space. Within this retained subset, model families were compared using Kruskal–Wallis tests followed by Dwass–Steel–Critchlow–Fligner pairwise comparisons. These post-selection comparisons were used to characterize differences among model families within the high-performing parameter region and should not be interpreted as unconditional tests across the full parameter space. Statistical significance was assessed at $p<0.05$. The individual parameter configuration with the highest language-likeness score was selected for the subsequent dynamical analyses.

\subsubsection*{Completion-only control}

To assess the contribution of processing pressure beyond structural completion, we compared the selected DLT--surprisal model with a completion-only control. Both models used the same parameters and structural and contextual generation processes, including pressure-dependent resolution of open relations. They differed in how boundaries were generated once all relations had been resolved. In the pressure-dependent model, boundary probability depended on accumulated processing pressure. In the completion-only control, boundaries were emitted with a fixed probability, while continuation and the opening of new relations were sampled independently of pressure. Both models prohibited boundaries before a minimum segment length of three events and enforced closure after 99 events when no relations remained open.

The fixed termination probability was calibrated to match the mean segment length of the pressure-dependent model using 30 random seeds, with 120 segments per seed, through an 18-iteration bisection search. Both models were then evaluated across 500 random seeds, each generating 120 segments. We compared seed-level segment-length statistics, including mean, standard deviation, coefficient of variation, and skewness, alongside language-likeness scores and boundary-aligned profiles of processing demands. Statistical comparisons used paired $t$-tests across matched random seeds.

\subsubsection*{Boundary-aligned simulation dynamics}

The selected parameter configuration belonged to the combined DLT–surprisal model family. The complete parameter set is reported in Supplementary Table S3. To assess the stability of its behavior, the selected configuration was rerun across 500 random seeds.

For each simulation, emitted boundary markers were identified from the model output. Because a boundary marker occurred immediately after the final generated unit of a segment, the boundary center was defined as the index of the boundary marker minus one. Aligned position 0 therefore corresponded to the final pre-boundary unit, and position +1 corresponded to the first post-boundary unit. The boundary crossing was located between positions 0 and +1. Boundary-aligned windows extended from six positions before to six positions after the boundary center. Boundary centers for which a complete alignment window was unavailable were excluded.

Within each window, we extracted total processing pressure, normalized boundary probability, contextual surprisal, memory cost, integration cost, and semantic mismatch. Semantic mismatch was defined as the complement of the similarity between the current token embedding and the model’s context state. 

For each variable, boundary-aligned profiles were first averaged across all eligible boundaries within each simulation. Seed-level profiles were then averaged across the 500 simulations. This procedure prevented simulations containing more boundary events from contributing disproportionately to the group-level profile. Uncertainty was quantified by bootstrap resampling across random seeds.

Non-boundary controls were sampled from the same simulated sequences. Candidate control centers were required to have a complete alignment window and to be more than three positions away from any true boundary center. For each simulation, control centers were randomly sampled to match the number of true boundary centers; sampling with replacement was used only when the number of eligible control positions was smaller than the number of true boundaries. The same variables and alignment window were then extracted around control centers and averaged for comparison.

\subsubsection*{Timescale estimation at multiple levels}
We analyzed the temporal scales generated by the selected pressure-constrained model across 500 random seeds. For continuous model variables, including boundary probability, memory cost, integration cost, and contextual surprisal, we estimated the dominant period using Welch spectral analysis. For each seed and variable, the power spectrum was computed after mean-centering the time series. The dominant period was defined as the inverse of the non-zero frequency with maximal spectral power. For discrete structural events, we estimated timescales from event intervals rather than spectral peaks. Relation openings, relation closures, and emitted boundaries were converted into binary event series, and the characteristic scale was defined as the median inter-event interval within each seed. Relation span was computed as the distance between the opening and closure of a resolved relation. Seed-level estimates were averaged across the 500 simulations, with uncertainty summarized using bootstrap confidence intervals. Full spectral settings and event-scale definitions are provided in the SI.

\subsection*{Human perceptual chunking: dataset and boundary choices}

We reanalyzed the publicly available perceptual chunking dataset \cite{vetchinnikova_chunking_2023}. In the original experiment, 50 linguistically naive participants listened to 97 short extracts of natural speech and marked perceived chunk boundaries in real time using a tablet-based application. The dataset provides individual boundary judgments aligned to word positions, together with annotations of grammatical structure, prosodic boundary strength, and pauses. After excluding extract-final positions, our analysis included 5,140 candidate word positions. For each position, boundary choices were aggregated across the 50 listeners into the number of boundary selections and non-selections.

\subsubsection*{Dependency-based memory and integration costs}

Structural processing costs were derived from dependency parses generated by the English parser of Stanza \cite{qi_stanza_2020}. For the chunking dataset, parsing was performed separately within the top-level syntactic units identified in the original annotations. We quantified two complementary costs following the distinction in dependency locality theory (DLT): memory cost, reflecting unresolved dependency relations, and integration cost, reflecting the resolution of dependencies across intervening material.

Memory cost was defined as the number of dependency arcs remaining open after each word. An arc connecting positions $i$ and $j$ was treated as active between its leftmost and rightmost endpoints. Post-word memory cost was calculated as
\begin{equation*}
M_t = \sum_{(i,j)\in\mathcal{A}}
\mathbb{1}\left[\min(i,j)\leq t<\max(i,j)\right],
\end{equation*}
where $\mathcal{A}$ denotes the set of dependency arcs within the parsing unit.

Integration cost was assigned when a dependency relation was resolved at its rightmost endpoint. For each resolved relation, we counted the intervening content-bearing discourse referents, operationalized as nouns, proper nouns, and verbs. For lexical dependencies, the current content-bearing word was additionally counted as a newly introduced referent, while function-word dependencies did not receive this additional cost. Heuristic antecedent links were also included for relative pronouns where applicable. Integration costs from multiple relations resolved at the same word were summed:
\begin{equation*}
I_t = \sum_{r\in\mathcal{R}_t}d_r,
\end{equation*}
where $\mathcal{R}_t$ denotes relations resolved at position $t$ and $d_r$ their corresponding integration costs.

\subsubsection*{Local semantic mismatch}

Lexical-semantic mismatch was estimated using 300-dimensional fastText word embeddings. For each word $w_t$, a local context vector was constructed by averaging the embeddings of the preceding $k$ words. Semantic mismatch was defined as the cosine distance between the current word representation and its local context:
\begin{equation*}
D_t^{(k)} =
1-\cos\left(v_{w_t},\bar{v}_{t-k:t-1}\right).
\end{equation*}
Higher values indicate greater semantic dissimilarity between the incoming word and the preceding context. The primary analysis used a window of four preceding words ($k=4$). Context was maintained within each speech extract without resetting at syntactic unit boundaries. Unlike the simulation, which quantified contextual demands using surprisal, the human analyses used local semantic mismatch as a complementary measure of contextual processing.

\subsubsection*{Statistical analysis of boundary choices}

We fitted grouped binomial logistic regression models to predict boundary choices at each candidate word position. Let $Y_{et}$ denote the number of listeners selecting a boundary at position $t$ in extract $e$, with $Y_{et}\sim\mathrm{Binomial}(n_{et},\pi_{et})$, where $n_{et}$ is the number of listener choices and $\pi_{et}$ is the probability of boundary selection.

The baseline model included grammatical boundary strength, prosodic boundary strength, normalized word position within the extract, and extract fixed effects. The full model additionally included post-word memory cost, integration cost, and local semantic mismatch:
\begin{equation*}
\begin{aligned}
\mathrm{logit}(\pi_{et}) =\;&
\beta_0+\alpha_e+\beta_1 P_{et}
+\beta_2 G_{et}+\beta_3 R_{et}\\
&+\beta_4 M_{et}
+\beta_5 I_{et}
+\beta_6 D_{et},
\end{aligned}
\end{equation*}
where $P$ denotes normalized word position, $G$ grammatical boundary strength, $R$ prosodic boundary strength, $M$ post-word memory cost, $I$ integration cost, $D$ semantic mismatch, and $\alpha_e$ extract fixed effects. The three processing-cost predictors were z-standardized across candidate positions. The baseline model omitted $M$, $I$, and $D$.

To examine the contribution of recent memory history, we extended the full model with memory cost at the immediately preceding word and the mean memory cost across the preceding three words:
\begin{equation*}
\mathrm{logit}(\pi_{et}) =
\eta_{et}^{\mathrm{full}}
+\beta_7 M_{e,t-1}
+\beta_8 \overline{M}_{e,t-3:t-1},
\end{equation*}
where $\eta_{et}^{\mathrm{full}}$ denotes the linear predictor of the full model. Memory-history measures were computed within individual syntactic parsing units and z-standardized within the analysis sample. Positions without a complete three-word history and annotated top-level unit endings were excluded, leaving 3,660 candidate positions.

We performed two sensitivity analyses. First, we excluded 300 annotated top-level unit endings and repeated the primary analysis on the remaining 4,840 positions. Second, we replaced the numeric grammatical-strength predictor with a categorical specification. Model fit was compared using the Akaike information criterion (AIC), and the joint contributions of added predictors were evaluated using Wald tests. Standard errors were clustered by speech extract, and individual associations were reported as odds ratios (ORs).

\subsection*{Naturalistic Stories corpus: reading time analysis}

We analyzed word-level self-paced reading time from the Naturalistic Stories corpus, in which participants read narrative texts word by word and reading time was recorded at each word \cite{futrell_natural_2021}. Within each participant and each story, we replaced outliers in reading time (outside the range of 100-3000 ms \cite{futrell_natural_2021}) by the median value, following \cite{lo_periodic_2023}. Reading time was then averaged across participants and log-transformed. To reduce spillover from the preceding word, we residualized current log reading time using an ordinary least-squares model that included previous-word log reading time and story identity as predictors \cite{vasishth2006proper}:
\begin{equation*}
\log RT_t = \beta_0 + \beta_1 \log RT_{t-1} + C(\mathrm{item}) + \epsilon_t .
\end{equation*}
The residuals from this model were used as the spillover-adjusted reading-time measure in subsequent analyses.

\subsubsection*{Sentence-boundary coding}
Sentence boundaries were defined at the word level. A word was coded as a boundary word when it was the final lexical word of a sentence, and the following lexical word was coded as the first post-boundary word. Punctuation was used to identify sentence segmentation but was not treated as an independent word-level observation. Sentence-final status was obtained from the sentence segmentation of the Naturalistic Stories corpus and was kept identical across the reading-time, dependency-based, semantic, and LLM analyses.

For LLM analyses, each story was processed as a continuous sequence without resetting model context at sentence boundaries. Consequently, temporal trajectory measures at a sentence-final word incorporated both the incoming transition from the preceding word and the outgoing transition toward the first word of the following sentence. Across-layer measures, by contrast, described the transformation of the same word across model layers and did not incorporate the following word.

\subsubsection*{Processing cost estimation}
Memory cost, integration cost, and local semantic mismatch were computed following the procedures described for the human chunking analysis. Dependency-based costs were derived from sentence-level Stanza parses, while semantic mismatch was calculated using a four-word context window within each story. All measures were aligned with the word-level reading-time data.

\subsubsection*{LLM processing dynamics}
We used a Qwen3 model with 4 billion parameters (Qwen/Qwen3-4B) to derive word-level prediction and representation measures. The model was evaluated without parameter updating, with dropout disabled and gradients turned off. Each story was entered into the model as a continuous text. The model tokenizer occasionally divided corpus words into multiple subword tokens. Tokenizer offset mappings were used to align model tokens with the original corpus word spans. Whereas the speech-chunking transcripts contained no sentence punctuation, the LLM input retained the original punctuation of the written narratives to match the word-level input presented to human readers. Tokens encoding punctuation within a corpus word’s character span were assigned to that word and included in its word-level measures. Word-level surprisal was obtained by summing the surprisals of all assigned tokens, and word-level hidden states were obtained by averaging their representations separately at each model layer. For temporal measures based on distances or directions between words, each hidden-state dimension was standardized across word positions within each story before geometric quantities were calculated. For word-to-word geometric measures, hidden-state dimensions were z-scored within each story. Cosine updates were computed from L2-normalized word representations, while temporal curvature was calculated from the directions of successive representational changes. Local intrinsic dimensionality was estimated from the original last-layer hidden states. Layer curvature and length-to-chord ratio were computed from the original word-level representations across layers, without additional normalization of the hidden-state vectors.

\subsubsection*{Contextual surprisal}
For each token $w_t$, contextual surprisal was defined as the negative log probability assigned by the model to the word given its preceding context. When a word was split into multiple subword tokens $w_{1}, \ldots, w_{m}$ word-level surprisal was computed as the sum of subword surprisals:
\begin{equation*}
\mathrm{Surprisal}=-\sum_{j=1}^{m}
\log p(w_{j} \mid w_{<j}).
\end{equation*}
Higher values indicate lower model probability, and therefore greater contextual prediction difficulty.

\subsubsection*{Cosine update}
To characterize representational dynamics, we extracted word-level hidden states from the model. Let $h_t$ denote the hidden-state representation of word $w_t$. For measures based on changes across words, hidden-state dimensions were z-scored within each story before computing distances or angles, so that measures were not dominated by story-specific differences in representation scale. Cosine update was defined as the cosine dissimilarity between consecutive word representations:
\begin{equation*}
\mathrm{Update}_t = 1 - \frac{h_t \cdot h_{t-1}} {\lVert h_t\rVert\,\lVert h_{t-1}\rVert}
\end{equation*}
Higher values indicate larger representational changes from the preceding word.

\subsubsection*{Local intrinsic dimensionality}
Local intrinsic dimensionality (ID) estimated the local complexity of the hidden-state space around each word. Pointwise local ID was estimated from the last-layer word representations using the TwoNN algorithm with Python package scikit-dimension. For each word, a local neighborhood containing the 100 nearest word representations was identified within the same story, and intrinsic dimensionality was estimated from the relative nearest-neighbor distances within that neighborhood. Words for which a complete neighborhood could not be defined were excluded. Higher values indicate that the local representation occupies a higher-dimensional and less compressible neighborhood.

\subsubsection*{Trajectory curvature}
Trajectory curvature measured how much the model's representational trajectory changed direction as the story unfolded word by word. This measure was computed from last-layer word representations. Let $h_t$ denote the hidden-state representation of word $w_t$, after averaging over all subword tokens aligned to that word. For word-to-word geometry, hidden-state dimensions were z-scored within each story before computing trajectory measures.

For each word $w_t$, we first computed two adjacent step vectors in hidden-state space. The incoming step captured the representational change from the previous word to the current word,
\begin{equation*}
\Delta^-_t = h_t - h_{t-1},
\end{equation*}
and the outgoing step captured the representational change from the current word to the next word,
\begin{equation*}
\Delta^+_t = h_{t+1} - h_t.
\end{equation*}
Trajectory curvature was then defined as the cosine distance between these two step vectors:
\begin{equation*}
\mathrm{Trajectory\_curvature}_t=
1 -
\frac{
\Delta^-_t \cdot \Delta^+_t
}{
\lVert \Delta^-_t \rVert
\lVert \Delta^+_t \rVert
}.
\end{equation*}
Low values indicate that the trajectory continued in a similar direction across $w_t$, whereas high values indicate a sharper turn in representational space.

\subsubsection*{Layer curvature}
Trajectory curvature characterizes changes in representational direction across successive words. We additionally quantified how the representation of a single word changed as it passed through the internal layers of the model. Following the layer-trajectory approach used in prior work on LLM representational curvature \cite{sipio_curved_2025}, for each word, hidden states from successive transformer layers were treated as an across-layer representational trajectory. The input embedding state was excluded, and only transformer-layer outputs were included.

Let $h_t^{(\ell)}$ denote the hidden-state representation of word $w_t$ at layer $\ell$, after averaging over all subword tokens aligned to that word. For a model with $L$ layers, each word therefore has a layer-wise trajectory $h_t^{(1)}, h_t^{(2)}, \ldots, h_t^{(L)}$. We first computed the step vector between consecutive layers:
\begin{equation*}
\delta_t^{(\ell)}=h_t^{(\ell+1)} - h_t^{(\ell)},
\end{equation*}
which describes how the representation of word $w_t$ changes from layer $\ell$ to layer $\ell+1$.

Layer curvature was then computed by comparing the directions of adjacent layer steps. For each intermediate layer, we calculated the cosine distance between the incoming layer step and the outgoing layer step:
\begin{equation*}
\kappa_t^{(\ell)}
=
1 -
\frac{
\delta_t^{(\ell-1)} \cdot \delta_t^{(\ell)}
}{
\lVert \delta_t^{(\ell-1)} \rVert
\lVert \delta_t^{(\ell)} \rVert
},
\qquad \ell = 2,\ldots,L-1 .
\end{equation*}
This value is low when the representation continues to move in a similar direction across adjacent layers, and high when the layer-wise transformation changes direction sharply. We summarized this local turning by averaging across intermediate layers:
\begin{equation*}
\mathrm{Layer\_curvature}_t
=
\frac{1}{L-2}
\sum_{\ell=2}^{L-1}
\kappa_t^{(\ell)} .
\end{equation*}
Mean layer curvature therefore captures the average local bending of a word's representation as it is transformed through the model hierarchy.

We also computed the layer length-to-chord ratio (LCR), which captures the global indirectness of the same across-layer trajectory. The layer path length was defined as the total distance traveled across consecutive layers:
\begin{equation*}
P_t
=
\sum_{\ell=1}^{L-1}
\left\lVert
h_t^{(\ell+1)} - h_t^{(\ell)}
\right\rVert .
\end{equation*}
The chord length was defined as the direct distance between the first and final layer representations:
\begin{equation*}
C_t
=
\left\lVert
h_t^{(L)} - h_t^{(1)}
\right\rVert .
\end{equation*}
The length-to-chord ratio was then 
\begin{equation*}
\mathrm{LCR}_t=\frac{P_t}{C_t}.
\end{equation*}

A value close to 1 indicates that the representation moves along an approximately straight path from lower to higher layers. Larger values indicate that the across-layer transformation follows a more indirect or curved path. Thus, mean layer curvature and LCR provide complementary descriptions of within-token geometry: mean layer curvature measures local turning between adjacent layer steps, whereas LCR measures the global deviation of the full layer trajectory from a straight line.

\subsubsection*{Regression analysis of pressure-derived predictors and boundary control}

To test whether pressure-derived variables explained behavioral and LLM outcomes beyond boundary-aligned visualization, we fitted regression models at the word level. The predictors of interest were memory cost, integration cost, and static semantic mismatch. Outcomes included spillover-adjusted reading-time residuals, Qwen surprisal, cosine update, local intrinsic dimensionality, trajectory curvature, mean layer curvature, and layer length-to-chord ratio.

All continuous predictors and outcomes were standardized within story. For each outcome $Y_t$, we first fitted a model without explicit sentence-boundary control:
\begin{equation*}
Y_{t,z}=\beta_0+\beta_1 M_{t,z}+\beta_2 I_{t,z}+\beta_3 D_{t,z}+\beta_4 P_{t,z}+C(\mathrm{item})+\epsilon_t,
\end{equation*}
where $M_t$ is memory cost, $I_t$ is integration cost, $D_t$ is static semantic mismatch, and $P_t$ is normalized word position within the story. Story identity was included as a fixed effect. We then fitted a second model including explicit sentence-boundary status:
\begin{equation*}
Y_{t,z}=\beta_0+\beta_1 M_{t,z}+\beta_2 I_{t,z}+\beta_3 D_{t,z}+\beta_4 P_{t,z}+C(\mathrm{item})+C(B_t)+\epsilon_t ,
\end{equation*}
where $B_t$ indicates whether word $w_t$ occurred at a sentence boundary.

Models were fitted using ordinary least squares with cluster-robust standard errors at the sentence level. For each predictor, we reported Wald (z)-statistics and FDR-corrected significance across coefficient tests. To quantify the contribution of explicit boundary status, we compared the two models using
\begin{equation*}
\Delta \mathrm{AIC}
=
\mathrm{AIC}_{\mathrm{with\ boundary}}
-
\mathrm{AIC}_{\mathrm{without\ boundary}}.
\end{equation*}

Negative values indicate that adding sentence-boundary status improved model fit. This comparison allowed us to distinguish graded pressure-derived effects from the additional contribution of discrete sentence-boundary position.

\section*{Data and code availability}
The perceptual speech-chunking data are publicly available through the Chunk Boundary Predictors dataset (\url{https://osf.io/7bta5/}). The narrative texts and self-paced reading data are available from the Natural Stories corpus (\url{https://github.com/languageMIT/naturalstories}). All analysis scripts and simulation
data are available at \url{https://github.com/RuiHe1999/boundary_simulation_RTLLM}.

\section*{Acknowledgements}
This study was funded by European Research Council (ERC-2023-SyG, 101118756 to WH, IS, and PH). The views and opinions expressed are those of the author(s) only and do not necessarily reflect those of the European Union.

\section*{References}
\bibliography{zHenriquesLab-Mendeley}

\captionsetup*{format=largeformat}
\section*{Simulation Model}

\subsection*{Model overview}

We implemented a stochastic generative simulation to examine whether boundary-like organization can emerge from local constraints on sequential processing. The model generates a stream of discrete events and lexical tokens. Crucially, sentence boundaries are not supplied to the model. Instead, boundaries are generated endogenously as a consequence of accumulated processing pressure from structural dependency and contextual compatibility. The simulation is based on the assumption that linguistic boundaries are not arbitrary delimiters, but tend to occur at positions where processing demands accumulate or where previously processed material must be integrated. This is consistent with the wrap-up effects at sentence final \cite{stowe2018sentence}, in which boundary positions are associated with increased processing time and have been interpreted as sites of syntactic, semantic, or discourse-level integration. 

We model this processing pressure as arising from two complementary sources \cite{rathi-2021-dependency}. First, dependency locality theory (DLT) proposes that sentence processing difficulty depends on the cost of maintaining incomplete dependencies in memory and integrating syntactically related elements across distance. Second, surprisal-based accounts propose that processing difficulty increases when an incoming word is less expected given the preceding context. The present simulation combines these two sources of difficulty in a stochastic generative model, asking whether boundary-like structures can emerge from the accumulation and release of structural and lexical-informational pressure.

This simulation is not intended to reproduce the full grammatical structure of natural language. Rather, it formalizes a minimal mechanistic hypothesis: boundary-like segmentation can emerge from the accumulation and release of processing pressure in a sequential system.

\subsection*{Event alphabet} 

At each position $t$, the model emits one event ($e_t$) from a four-symbol event alphabet:

\begin{equation}
e_t \in \{X, O, C, |\}.
\end{equation}

The four event types are defined as follows:

\begin{align*}
X & : \text{ordinary lexical event}, \\
O & : \text{opening of a dependency-like relation}, \\
C & : \text{closure of a dependency-like relation}, \\
| & : \text{boundary event}.
\end{align*}

The events (X), (O), and (C) emit lexical tokens. The boundary event (|) does not emit a lexical token. Instead, it terminates the current segment, resets the structural state, resets processing pressure, and initializes a new semantic context.

The model tracks the set of referential positions generated within the current segment, $\mathcal{R}_t$, and a stack of currently open dependency-like relations, $\mathcal{O}_t$. An opening event ($O$) pushes the current position onto the stack, whereas a closing event ($C$) pops the most recent opening from the stack. This last-in-first-out mechanism allows nested dependencies, because other events, including additional openings, may occur before an earlier opening is closed. However, it rules out crossing dependencies. A boundary event is legal only when the stack is empty.

A boundary is structurally legal only when no relation remains open:

\begin{equation}
|\mathcal{O}_t| = 0.
\end{equation}

If at least one relation is still open, the model cannot generate a boundary. This constraint prevents the system from terminating a segment before the unresolved structural material has been integrated.

\subsection*{Synthetic lexical-semantic embedding system}

At each position $t$, the model selects a token from a synthetic lexical-semantic space. The vocabulary contains 200 token types, and each word $w$ is represented by a 32-dimensional embedding vector $\mathbf{v}_w$. The lexical space is generated from $K=8$ latent topic centers. Each topic center $\mathbf{c}_k$ is sampled from a Gaussian distribution and normalized to unit length. Each word is assigned to one latent topic $z_w$, and its embedding is generated by adding Gaussian noise to the corresponding topic center:

\begin{equation}
\mathbf{v}_w =
\frac{\mathbf{c}_{z_w} + \boldsymbol{\epsilon}_w}
{\|\mathbf{c}_{z_w} + \boldsymbol{\epsilon}_w\|_2},
\quad
\boldsymbol{\epsilon}_w \sim \mathcal{N}(\mathbf{0}, \sigma_{\mathrm{emb}}^2 \mathbf{I}).
\end{equation}

This creates a weak semantic geometry, where words from the same topic tend to be more similar than words from different topics. We did not generate 200 word embeddings as independent random vectors, because random vectors in high-dimensional spaces tend to be approximately orthogonal to each other \cite{kanerva_hyperdimensional_2009}, which would eliminate the local similarity structure expected in a lexical-semantic space \cite{turney2010frequency}.

\subsection*{Structural costs: memory and integration}

As inspired by the dependency locality theory \cite{rathi-2021-dependency}, structural costs include both memory and integration. Memory cost is defined as the number of currently open relations after the current event has been processed:
\begin{equation}
M_t = |\mathcal{O}_t|.   
\end{equation}

This term captures the burden of maintaining unresolved structural relations during sequential processing. More open relations imply greater memory pressure.

Integration cost is incurred when the model closes a relation. Suppose that a relation was opened at position $j$ and closed at position $t$. The integration cost is defined as the number of referential positions between the opening and closing positions:

\begin{equation}
I_t =
\sum_{r \in \mathcal{R}_t}
\mathbf{1}(j < r < t).
\end{equation}

For ordinary events (X) and opening events (O), integration cost is zero. Each non-boundary event also carries a unit discourse cost of 1, $$D_t=1$$, while boundary events incurred no discourse cost. The structural contribution to processing pressure therefore included both this unit cost and the distance-based integration cost $$I_t$$. 

Thus, for a non-boundary event, the structural contribution to pressure depends on the current memory load and on the cost of integrating a relation when a closure occurs.

\subsection*{Surprisal costs: contextual surprisal}

Contextual surprisal was derived from the probability of a token given its preceding semantic context. Every segment starts with setting up a local semantic context. At a non-boundary position $t$, compatibility between a token vector \(v_w\) and the current context \(m_t\) is computed using cosine similarity:
\begin{equation}
c(w,t) = v_w^\top m_{t}.
\end{equation}

The semantic context is not based on a fixed window. Instead, it is continuously updated using exponentially decaying memory of previous tokens:
\begin{equation}
m_{t}=(1-\eta)m_{t-1}+\eta v_{w_{t-1}},
\end{equation}

and then normalized as:
\begin{equation}
\mathbf{m}_{t}
=
\frac{\tilde{\mathbf{m}}_{t}}{\|\tilde{\mathbf{m}}_{t}\|_2}.
\end{equation}

The parameter $\eta$ controls the context update rate. In the present simulations, $\eta = 0.15$. Semantic context changes gradually within a segment and is reset after a boundary.

The probability of sampling token $w$ is then given by a softmax over contextual compatibility:
\begin{equation}
p(w_t \mid \mathbf{m}_{t}) = \frac{
\exp{\beta_{\mathrm{ctx}} c(w,t)}
}{
\sum_{u=1}^{V} \exp{\beta_{\mathrm{ctx}} c(u,t)}
}.
\end{equation}

The parameter $\beta_{\mathrm{ctx}}$ controls the strength of the semantic contextual constraint. When $\beta_{\mathrm{ctx}}=0$, the lexical distribution is approximately uniform. As $\beta_{\mathrm{ctx}}$ increases, words that are more compatible with the current segment context become more likely. After a word is sampled, its lexical surprisal is computed as the negative log-likelihood conditioned by the context:
\begin{equation}
S_t = -\log p(w_t \mid \mathbf{m}_{t}).
\end{equation}

\subsection*{Processing pressure}
The central dynamic variable of the simulation is processing pressure, denoted by $P_t$. For non-boundary events, pressure is updated as a non-negative weighted sum from structural and surprisal costs. 

\begin{equation}
P_t =
\max
\left(
0,
\rho P_{t-1}
+
a_{\mathrm{mem}} M_t
+
b_{\mathrm{int}}(D_t + I_t)
+
\gamma_{\mathrm{surp}} S_t
+
\epsilon_t
\right),
\quad
\epsilon_t \sim \mathcal{N}(0,\sigma_{\epsilon}^{2}).
\end{equation}

Rectification by $\max({0,\cdot})$ prevents pressure from becoming negative. $\rho$ controls the persistence of pressure across events. $a_{\mathrm{mem}}$, $b_{\mathrm{int}}$, and $\gamma_{\mathrm{surp}}$ are, respectively, the weights for memory, integration, and surprisal costs. Although $\beta_{\mathrm{ctx}}$ does not directly add a separate term to processing pressure, it shapes the contextual probability distribution, and therefore affects pressure indirectly through surprisal $S_t$. $\sigma_{\epsilon}$ is the standard deviation of pressure noise and set as 0.05 in the present simulations.

Pressure is reset at the boundary event to zero. In other words, pressure accumulates during non-boundary processing and gets released at boundaries.

\subsection*{Boundary drive}
Processing pressure is transformed into boundary drive $q_t$ using a sigmoid function:
\begin{equation}
q_t = \sigma{k(P_{t-1}-\theta)}, 
\end{equation}

where the sigmoid function is defined as:
\begin{equation}
\sigma(x) = \frac{1}{1+\exp(-x)}. 
\end{equation}

In this drive function, $k$ controls the steepness of the mapping from pressure to boundary drive and $\theta$ is the pressure threshold. When pressure is far below threshold, $q_t$ is close to zero. When pressure exceeds threshold, $q_t$ increases, making a boundary or closure event more likely depending on the current structural state.

\subsection*{Event emission}
The event-emission rule depends on whether the model currently has unresolved open relations. At each position, the model first computes a pressure-derived drive from the current pressure state. It then samples one legal event according to the current structural state. If the sampled event is non-boundary, the model samples a lexical token, computes memory, integration, and surprisal costs, updates pressure, and updates the semantic context. If the sampled event is a boundary, no lexical token is emitted; pressure, open relations, referents, segment length, and semantic context are reset.

\subsubsection*{No open relation}
When no relation is open, a boundary is structurally legal. The model defines unnormalized event weights for ordinary events, opening events, and boundary events:
$u_X=1, u_O= b_{\mathrm{open}}(1-q_t), u_{|}=q_t$. The closure weight is zero $u_C = 0$, because there is no open relation to close. The event probabilities are obtained by normalizing the legal event weights: 
\begin{equation}
p(e_t=e) =
\frac{u_e}{\sum_{g \in \mathcal{E}_t} u_g}.
\end{equation}

where $\mathcal{E}_t$ is the set of legal events at position $t$.

In plain terms, when pressure is low, the model is more likely to continue the segment or open a new relation. When pressure is high and no relation remains open, the boundary weight increases, making a boundary more likely.

The model also imposes minimum and maximum segment-length constraints. If the current segment length is below the minimum segment length (set as 3), the boundary weight is forced to zero. If the current segment length reaches the maximum segment length (set as 99), the boundary weight is forced to one. 

\subsubsection*{At least one open relation}
When at least one relation is open, a boundary is illegal. In this state, pressure promotes closure rather than boundary insertion. 

When the number of opening relations is below the maximum allowed number (set as 6), the opening and closing weights are:
\begin{equation}
u_C =
base_{\mathrm{close}}
+
(1-base_{\mathrm{close}})q_t;\quad
u_O =
base_{\mathrm{open}}(1-q_t).
\end{equation}

If the maximum number of open relations has been reached, opening is disallowed:
\begin{equation}
u_C =
base_{\mathrm{close}}
+
(1-base_{\mathrm{close}})q_t;\quad
u_O = 0.
\end{equation}

This emission system creates a pressure-sensitive structural cycle. When pressure is low, the system can continue the segment and open relations. When pressure rises while relations remain open, the system tends to close them. Once all relations are closed, high pressure increases the probability of a boundary. The boundary then resets pressure and starts a new segment.

\subsection*{Boundary as pressure release}
A boundary event is treated as a transition in the model dynamics, not as an input label or externally supplied delimiter. At each step, segmentation is possible only when the current structural state can be closed, and its probability increases with accumulated processing pressure. When such an event occurs, it has four consequences. First, it terminates the current segment. Second, it resets the segment length to zero. Third, it clears all referents and open relations from the current structural state. Fourth, it resets processing pressure and initializes a new segment-level semantic context. Boundaries are therefore defined by the system's own pressure-release dynamics. Boundary probability increases with accumulated processing pressure when the structural configuration permits closure.

\subsection*{Parameter scan}
We performed a systematic parameter scan over the main pressure and boundary-generation parameters. The fixed architectural parameters are listed in Table~\ref{tab:fixed_parameters}. These parameters define the background simulation environment and were held constant across the parameter scan.

The manipulated parameters are summarized in Table~\ref{tab:tuned_parameters}. These parameters control boundary threshold, pressure persistence, structural pressure, lexical surprisal, contextual constraint, and the baseline tendency to open or close relations. The parameter grid was organized into three model families. In the DLT-only family, memory and integration weights were varied while lexical surprisal was removed from the pressure equation by setting $\gamma_{\mathrm{surp}}=0$. In the surprisal-only family, memory and integration costs were removed by setting ($a_{\mathrm{mem}} = 0$, $b_{\mathrm{int}} = 0$), while $\gamma_{\mathrm{surp}}$ was varied, including a zero-surprisal baseline. In the DLT-surprisal family, both DLT-inspired structural costs and lexical surprisal contributed to pressure.

Across all families, the scan contained 20,580 unique parameter combinations. Each parameter combination was first evaluated across ten random seeds ($7,14,21,28,35,42,49,56,63,70$). For each seed, the model generated 120 completed segments.

\section*{Parameter Selection and Language-Likeness Score}

\subsection*{Rationale}

The simulation is not a natural language model, and it is not intended to generate fully linguistic sequences. Its output could therefore be only evaluated at the level of broad statistical signatures that are characteristic of linguistic organization. We focused on two such signatures: a Zipf-like lexical rank-frequency distribution and a decay of dependency-like relations with linear distance \cite{ degiuli2019random, piantadosi2014zipf}. These statistics provide weak but useful constraints on whether a simulated corpus has non-degenerate language-like structure, without implying that the generated sequences are equivalent to natural language.

The remaining components served as quality-control criteria, excluding simulations with insufficient lexical diversity, excessive dominance by individual tokens, limited vocabulary coverage, or too few generated tokens. The resulting language-likeness score was therefore a heuristic ranking criterion: it identified parameter regimes that produced statistically plausible and analyzable sequences, rather than estimating a formal likelihood or model evidence.

Each simulated corpus corresponded to one parameter combination and one random seed. For each corpus, we extracted dependency-length statistics, lexical frequency statistics, and quality control based on lexical diversity statistics and generated sequence length.

\subsection*{Dependency-length decay}
The first component evaluated whether dependency-like links showed systematic length-frequency decay. Each closed relation defined a link between its opening position $j$ and its closing position $t$. The link length was defined as $\ell = t - j$. 

For each simulated corpus, we computed the empirical frequency distribution of link lengths, $f(\ell)$. Previous studies showed that the relations between $\ell$ and $f(\ell)$ could fit either an exponential model ($\log f(\ell) = c - \lambda \ell$) or a power-law model ($\log f(\ell) = c - \alpha \log \ell$). 

For each model, we retained the coefficient of determination, $R^2_{\mathrm{exp}}$, for the exponential model and $R^2_{\mathrm{power}}$ for the power-law model. These statistics quantified whether dependency-like links showed systematic distance decay.

The dependency decay score summarized whether dependency-like link lengths showed systematic distance decay. It was defined as the mean of the power-law and exponential $R^2$ values:

The resulting value was clipped to the interval from 0 to 1:

\begin{equation}
\mathrm{DepDecay}=\frac{
R^2_{\mathrm{power}} + R^2_{\mathrm{exp}}
}{2}, \quad
\mathrm{DepDecay}
\leftarrow
\min
\left(
1,
\max(0,\mathrm{DepDecay})
\right).
\end{equation}

\subsection*{Lexical distribution: Zipf's law}
Lexical frequency statistics were computed from the generated token sequence, excluding boundary events. Token types were ranked by frequency. Let $r$ denote token frequency rank and let $f(r)$ denote relative token frequency. We fitted a Zipf-like rank-frequency relation $\log f(r) = c - \alpha_{\mathrm{Zipf}}\log r$. This yielded a Zipf exponent $\alpha_{\mathrm{Zipf}}$ and a coefficient of determination $R^2_{\mathrm{Zipf}}$.

The Zipf score evaluated whether the generated lexical distribution showed a clear rank-frequency scaling relation and whether the estimated Zipf exponent was close to one \cite{piantadosi2014zipf}. For each simulated corpus, we fitted a linear model to the log-log rank-frequency relation and extracted both the coefficient of determination, \(R^2_{\mathrm{Zipf}}\), and the estimated exponent, \(\alpha_{\mathrm{Zipf}}\). We then defined an exponent score as

\begin{equation}
S_{\alpha}
=
\exp
\left(
-
\frac{|\alpha_{\mathrm{Zipf}} - 1|}{0.4}
\right).
\end{equation}

This score equals one when $\alpha_{\mathrm{Zipf}}=1$ and decreases exponentially as the exponent deviates from one.

\begin{equation}
S_{\mathrm{Zipf}}
=
R^2_{\mathrm{Zipf}} \cdot S_{\alpha}.
\end{equation}

Thus, a simulated corpus received a high Zipf score only when the rank-frequency relation was strong and the estimated exponent was close to the canonical Zipf value.

\subsection*{Anti-collapse quality control}
Four binary quality-control checks were conducted. They were used only as broad sanity checks, not as theoretical targets. Each criterion was coded as 1 if it was satisfied and 0 otherwise.

First, normalized lexical entropy was computed as
\begin{equation}
Q_{\mathrm{entropy}}
=
\mathbf{1}(H_{\mathrm{norm}} > 0.50).
\end{equation}

Second, the frequency share of the most common token was computed as
\begin{equation}
Q_{\mathrm{top1}}
=
\mathbf{1}(p_{\max} < 0.30).
\end{equation}

Third, vocabulary coverage was computed as
\begin{equation}
Q_{\mathrm{coverage}}
=
\mathbf{1}(\mathrm{Coverage} > 0.50).
\end{equation}

Fourth, the total number of generated lexical tokens was recorded as
\begin{equation}
Q_{\mathrm{tokens}}
=
\mathbf{1}(N_{\mathrm{tokens}} > 1000).
\end{equation}

The quality control score was defined as the conjunction of these four criteria:
\begin{equation}
Q = \mathbf{1}
\left(
Q_{\mathrm{entropy}} = 1,
Q_{\mathrm{top1}} = 1,
Q_{\mathrm{coverage}} = 1,
Q_{\mathrm{tokens}} = 1
\right).
\end{equation}

\subsection*{Composite language-likeness score}

The final language-likeness score was defined as the sum of three components:
\begin{equation}
\mathrm{DepDecay}
+
S_{\mathrm{Zipf}}
+
Q.
\end{equation}

The maximum possible value of the score is 3. A score close to 3 indicates that the simulated corpus simultaneously showed strong dependency-length decay, Zipf-like lexical rank-frequency structure, and passed all lexical quality-control criteria.

\subsection*{Aggregation across seeds}
The language-likeness score was first computed at the seed level. Scores were then averaged across seeds for each unique parameter combination. Parameter combinations were ranked by their mean language-likeness score across seeds. This aggregation reduced sensitivity to a single favorable stochastic realization. 

\subsection*{Parameter selection}
We first retained the top 10\% of parameter combinations ranked by mean language-likeness score. We then excluded any combination for which one or more seeds failed the quality-control criteria. This procedure retained a high-performing region of the parameter space rather than relying only on a single best parameter combination. The best parameter set was specifically picked up for later analysis. 

\section*{Simulation evaluation}
\subsection*{Model family comparison}
\begin{figure}
\centering
\includegraphics[width=\textwidth]{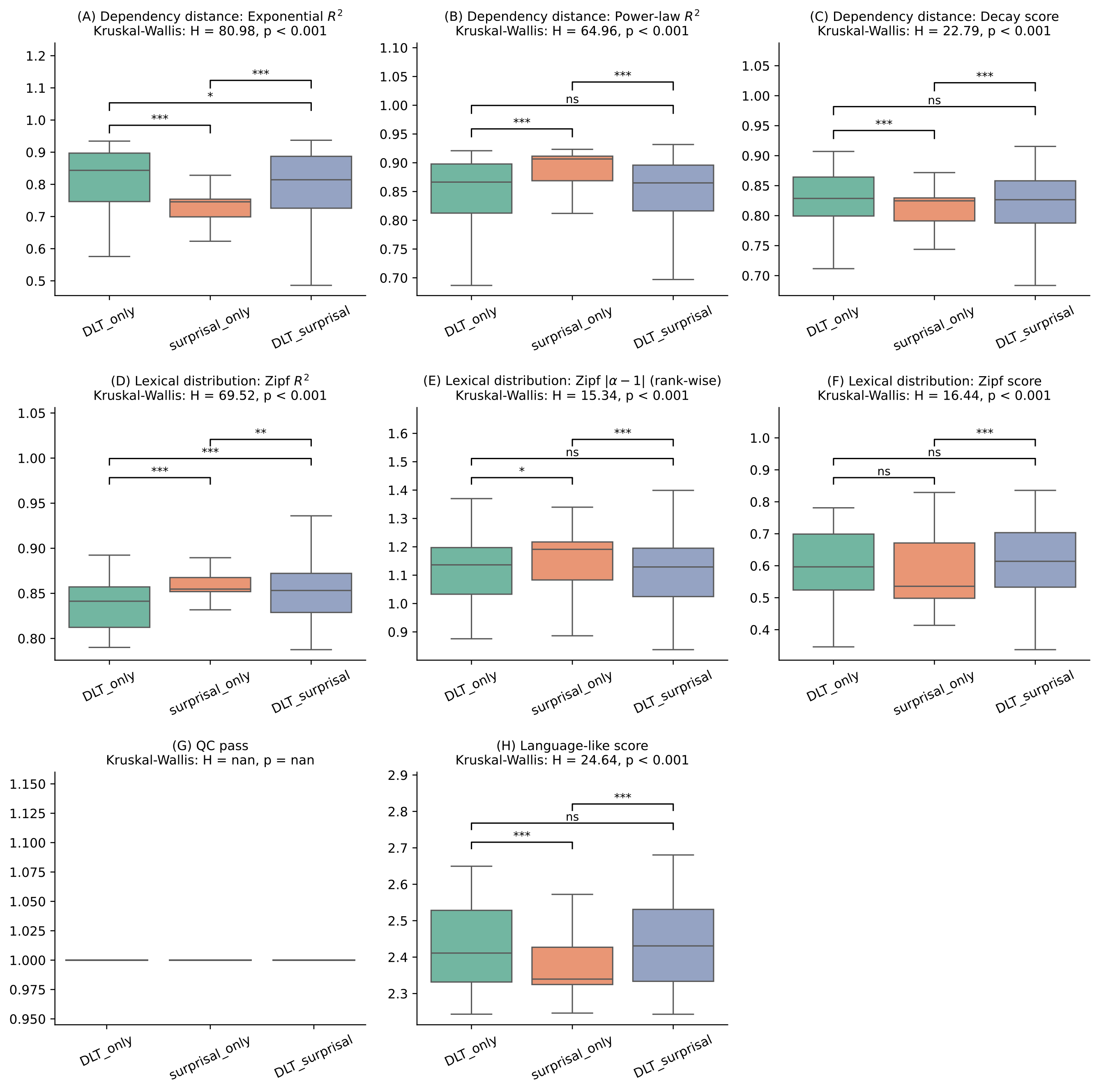}
\caption{Model-family differences in component language-likeness metrics.
Model families were compared among the retained top 10\% parameter regimes: DLT-only, surprisal-only, and combined DLT-surprisal models. DLT stands for dependency locality theory. Panels show group differences in dependency-distance exponential fit ($R^2$; A), dependency-distance power-law fit ($R^2$; B), dependency-distance decay score (C), lexical Zipf fit ($R^2$; D), deviation of the Zipf exponent from the ideal value of 1 (E), lexical Zipf score (F), quality-control pass rate (G), and the overall language-likeness score (H). Omnibus differences were tested using Kruskal--Wallis tests, followed by Dwass--Steel--Critchlow--Fligner pairwise comparisons. Significance brackets indicate pairwise comparisons: ns, not significant; * $p < 0.05$; ** $p < 0.01$; *** $p < 0.001$. Quality-control pass rate was constant across retained regimes and was therefore not tested.}
\label{fig:family_compare}
\end{figure}

Figure \ref{fig:family_compare} showed that different pressure components contributed to different aspects of language-like organization. Dependency-distance metrics were strongly affected by model family, including exponential fit, power-law fit, and the combined dependency-decay score. Models containing DLT-inspired structural costs outperformed the surprisal-only model on these dependency-based measures, indicating that structural pressure was the main driver of broad sequence-level organization. Lexical distribution metrics also differed across model families. Adding contextual surprisal to the structural model did not significantly improve the overall language-likeness score relative to the DLT-only model, but it improved the Zipf-related lexical profile, including the rank-frequency fit and Zipf-score components. Thus, structural costs primarily supported dependency-like organization and overall language-likeness, whereas surprisal contributed more specifically to the lexical-statistical structure of the generated sequences. For all measures, omnibus differences were tested using Kruskal–Wallis tests, followed by Dwass–Steel–Critchlow–Fligner pairwise comparisons. 

\subsection*{Parameter effects on language-like score}
\begin{figure}
\centering
\includegraphics[width=\textwidth]{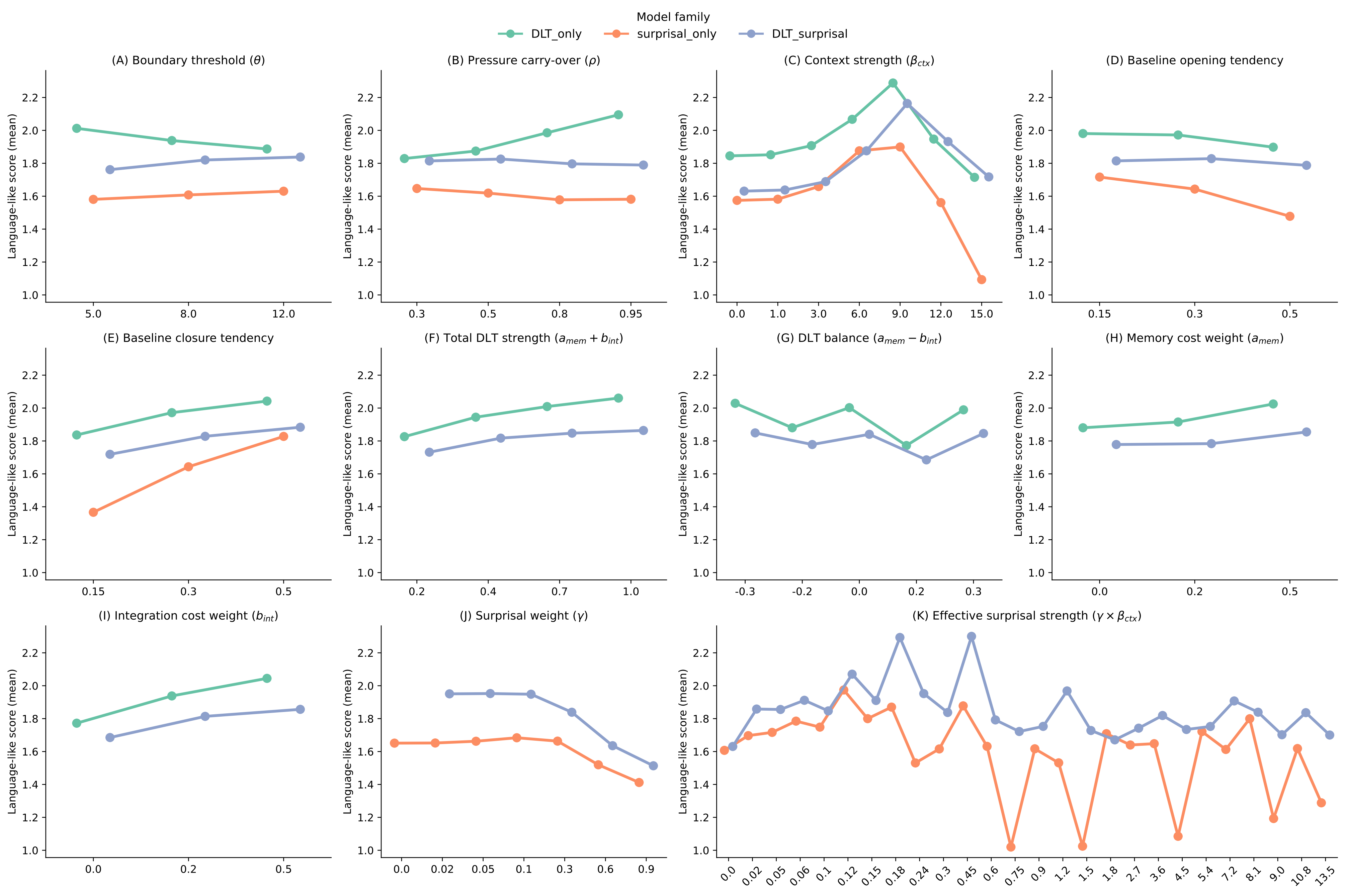}
\caption{
Parameter sensitivity of language-likeness across model families among the retained top 10\% parameter regimes. Mean language-likeness scores are shown as a function of individual simulation parameters, separately for DLT-only, surprisal-only, and combined DLT-plus-surprisal models. For each panel, scores were averaged over all other parameter settings within the scanned parameter space. Panels show (A) the boundary threshold $\theta$, (B) pressure carry-over $\rho$, (C) context strength $\beta_{\mathrm{ctx}}$, (D) baseline opening tendency, (E) baseline closure tendency, (F) total DLT strength $a_{\mathrm{mem}} + b_{\mathrm{int}}$, (G) DLT balance $a_{\mathrm{mem}} - b_{\mathrm{int}}$, (H) memory-cost weight $a_{\mathrm{mem}}$, (I) integration-cost weight $b_{\mathrm{int}}$, (J) surprisal weight $\gamma$, and (K) effective surprisal strength $\gamma \times \beta_{\mathrm{ctx}}$. The analyses show that language-like organization was not restricted to a single parameter setting. Models containing DLT-inspired structural costs generally achieved higher language-likeness scores than surprisal-only models across a broad range of thresholds, carry-over values, opening/closure tendencies, and structural-cost weights. Contextual parameters showed a more nonlinear profile, with strongest performance at intermediate context strengths and reduced scores when surprisal pressure became too dominant.
}
\label{fig:param_compare}
\end{figure}

We next used a descriptive marginal analysis to visualize how individual parameters were associated with language-likeness across the scanned parameter space. This analysis was not intended to estimate the independent causal effect of each parameter. Instead, it summarized which regions of the parameter space tended to produce more language-like simulated sequences. In Figure~\ref{fig:param_compare}, each point shows the mean language-likeness score for simulations sharing a given parameter value, computed separately for each model family and averaged over the remaining parameters.

Language-likeness was not controlled by a single parameter. The boundary threshold $\theta$ had only a modest effect, with scores remaining relatively stable across the tested threshold range. Pressure carry-over $\rho$ showed a clearer positive association, especially in the DLT-only family, indicating that some persistence of accumulated pressure helped generate more structured sequence organization rather than purely local boundary fluctuations.

Context strength $\beta_{\mathrm{ctx}}$ showed a nonlinear profile. Scores increased from weak to moderate context strength, but declined when contextual constraint became too strong, most clearly in the surprisal-only family. This suggests that contextual information was beneficial only within a bounded range: weak context provided little structure, whereas overly strong context could make lexical sampling too constrained and reduce the balance of the generated sequences.

Baseline opening and closure tendencies also shaped language-likeness. Opening tendency showed relatively weak effects in the DLT-containing families, whereas closure tendency showed a more consistently positive association. This pattern suggests that allowing relations to close at an appropriate rate helped maintain plausible dependency-like organization, preventing relations from remaining unresolved for too long.

The DLT-specific parameters showed the most stable associations with language-likeness. Higher total DLT strength, defined as $a_{\mathrm{mem}} + b_{\mathrm{int}}$, was associated with higher scores in both DLT-containing families. The separate memory and integration weights also showed positive associations, with integration cost $b_{\mathrm{int}}$ displaying the clearest monotonic increase. In contrast, the balance term $a_{\mathrm{mem}} - b_{\mathrm{int}}$ showed no simple monotonic trend, suggesting that performance depended more on the presence and overall strength of structural pressure than on a narrowly tuned dominance of memory over integration, or vice versa.

Surprisal-related parameters showed more selective effects. Increasing the direct surprisal weight $\gamma$ did not monotonically improve language-likeness and could reduce scores at high values. Effective surprisal strength, defined as $\gamma \times \beta_{\mathrm{ctx}}$, was also variable, indicating that lexical surprisal contributed to the final score only when appropriately scaled by contextual strength. 

\subsection*{Best parameter regime}
The best parameter regime, in terms of the highest language-likeness score, was presented in Table \ref{tab:best_parameter_regime}. The parameter set was rerun across 500 random seeds, from 0 to 499, for stability. This parameter regime was used for the timescale analysis as below. 

\subsection*{Timescale analysis of simulated dynamics}
For each seed, we extracted both continuous model variables and discrete structural events from the simulated sequence.

For continuous variables, we estimated the dominant period using Welch spectral analysis. The analyzed variables were boundary probability, integration cost, memory cost, and contextual surprisal. Integration cost was defined as the combined integration load,
\begin{equation}
I_t = I^\mathrm{disc}_t + I^{\mathrm{rel}}_t,
\end{equation}
where $I^{\mathrm{disc}}_t$ denotes the intervening-unit cost and $I^{\mathrm{rel}}*t$ denotes the resolved-relation cost. For each variable and seed, the time series was converted to a numeric array, missing values were removed, and the mean was subtracted. Power spectra were then estimated using Welch’s method with sampling frequency $f_s = 1$, segment length $n_\mathrm{perseg} = 512$, 50\% overlap, constant detrending, and density scaling. The zero-frequency component was removed. The dominant frequency was defined as the positive frequency at which the power spectral density reached its maximum:
\begin{equation}
\mathrm{PSD}(f^{*}) = \max_{f>0} \mathrm{PSD}(f).
\end{equation}

The corresponding timescale was then computed as:
\begin{equation}
T^{*} = \frac{1}{f^{*}}.
\end{equation}

This provided a seed-wise estimate of the dominant oscillatory scale for each continuous model variable.

For discrete structural events, we estimated characteristic scales using event intervals rather than spectral peaks. Relation opening events were defined as positions where the generated operation was O, relation closure events as positions where the operation was C, and actual boundaries as positions where a boundary marker was emitted. For each binary event series and seed, event positions were identified and the median inter-event interval was computed:
\begin{equation}
T_{\mathrm{event}} = \mathrm{median}\left(t_{i+1} - t_i\right).
\end{equation}

Seeds with fewer than two events of a given type were excluded for that variable. Relation span was computed separately as the distance between the opening and closure of a resolved relation,
\begin{equation}
L_{\mathrm{span}} = i_{\mathrm{close}} - i_{\mathrm{open}},
\end{equation}
and summarized by the median span within each seed.

All periods and scales were computed separately for each seed. Figure values show the mean across seeds, with 95\% bootstrap confidence intervals. Thus, continuous variables are reported as dominant Welch peak periods, whereas discrete structural quantities are reported as median inter-event intervals or median relation spans.

Specifically, we plotted the distribution of the median inter-boundary intervals (i.e. segment length) across all seeds, as in Figure~\ref{fig:seg_distribution}.

\section*{Completion-only control model}
To isolate the contribution of pressure-dependent boundary gating, we constructed a completion-only control using the same generative framework and parameters as the selected pressure-driven model (SI, Table S~\ref{tab:best_parameter_regime}). Both models generated sequences through four possible operations: emitting an ordinary token ($X$), opening a dependency-like relation ($O$), closing an existing relation ($C$), or emitting a segment boundary ($B$). Boundaries were permitted only when all open relations had been resolved. The models differed in how boundaries and subsequent continuation events were selected at these structurally permissible positions.

In the pressure-driven model, accumulated processing pressure $P_t$ was transformed into a gating signal through
\begin{equation}
q_t = \sigma\left[k(P_{t-1}-\theta)\right],
\end{equation}
where $k$ controls sigmoid steepness and $\theta$ the pressure threshold. When no relations remained open, the unnormalized weights for ordinary token emission, relation opening, and boundary emission were
\begin{equation}
w_X,w_O,w_B = 1,\;b_{\mathrm{open}}(1-q_t),\;q_t,
\end{equation}
where $b_{\mathrm{open}}$ is the baseline weight for relation opening. Event probabilities were obtained by normalizing these weights by their sum. Increasing pressure therefore simultaneously favored boundary emission and suppressed the opening of new relations.

In the completion-only model, this pressure-dependent choice was replaced by a fixed boundary hazard $h$. At positions with no open relations, the event probabilities were
\begin{equation}
\begin{aligned}
p(B) &= h,\\
p(X) &= \frac{1-h}{1+b_{\mathrm{open}}},\\
p(O) &= \frac{b_{\mathrm{open}}(1-h)}
             {1+b_{\mathrm{open}}}.
\end{aligned}
\end{equation}
Thus, once structural completion was reached, boundary emission and the relative probabilities of continuation events were independent of accumulated pressure.

Importantly, all processing-pressure dynamics preceding structural completion remained unchanged. When one or more relations were open, both models sampled ordinary token emission, relation opening, and relation closure from the same unnormalized weights:
\begin{equation}
w_X,w_O,w_C=
1,\;
b_{\mathrm{open}}(1-q_t),\;
b_{\mathrm{close}}+(1-b_{\mathrm{close}})q_t,
\end{equation}
with relation opening disabled when the maximum number of open relations was reached. Hence, accumulated pressure continued to promote the resolution of outstanding relations in both models. The memory, integration, and contextual surprisal contributions to pressure accumulation were identical, as were lexical generation and contextual updating. Both models also enforced a minimum segment length of three tokens and forced boundary emission at the maximum segment length of 99 tokens whenever no relations remained open. Boundary emission reset accumulated pressure to zero, cleared the local dependency and referent states, and initialized a new contextual state.

To control for differences in average segment length, the fixed hazard was calibrated against the pressure-driven model. The reference mean segment length was estimated from 30 random seeds, each generating 120 segments. A binary search over $h$ was then performed for 18 iterations using the same calibration seeds and segment count. This procedure yielded $h=0.13423$, corresponding to a target mean segment length of 22.07 tokens. Both models were subsequently evaluated over 500 paired random seeds, generating 120 segments per seed. We compared segment-length distributions, language-likeness scores, and boundary-aligned processing dynamics. Seed-level summary statistics were compared using paired $t$-tests.

\section*{Supplementary results}
\subsection*{Sentence length distribution}
\begin{figure}
\centering
\includegraphics[width=\textwidth]{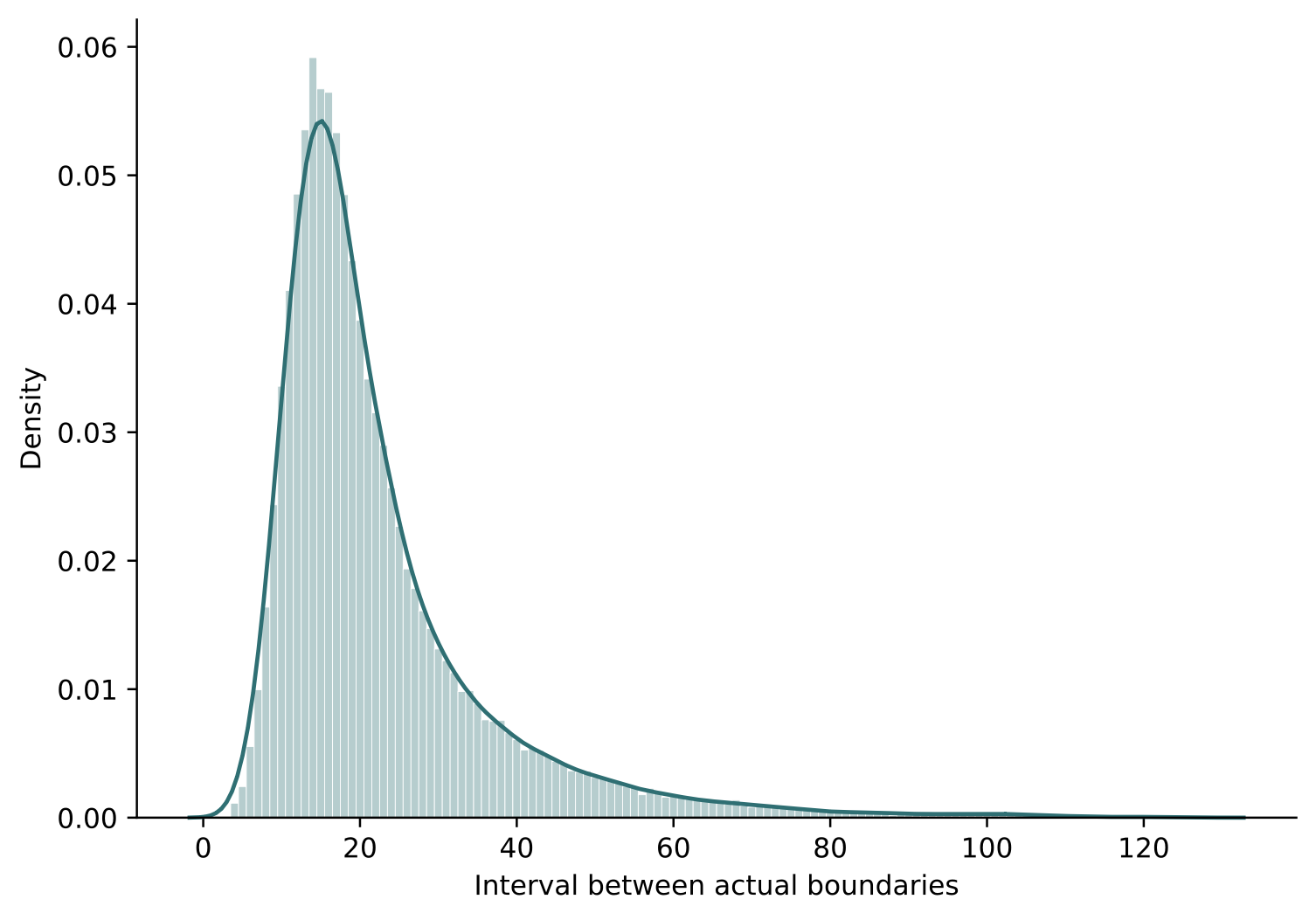}
\caption{
Actual sentence boundaries occur at a characteristic but variable timescale.
The plot shows the empirical distribution of token intervals between consecutive sentence boundaries. Most intervals clustered around short-to-intermediate spans, while the distribution showed a pronounced right tail, indicating that sentence lengths are variable rather than periodic. This empirical distribution was used as a reference for assessing whether emergent boundaries in the simulation occurred at naturalistic sentence-like intervals.
}
\label{fig:seg_distribution}
\end{figure}

The intervals between actual sentence boundaries showed a strongly right-skewed distribution with a long tail (Figure~\ref{fig:seg_distribution}). Most boundaries occurred after relatively short-to-intermediate token spans, with the highest density around the typical sentence-length range, whereas a smaller number of intervals extended far into a long right tail. 

\subsection*{Completion-only model comparison}

\begin{figure}
\centering
\includegraphics[width=\textwidth]{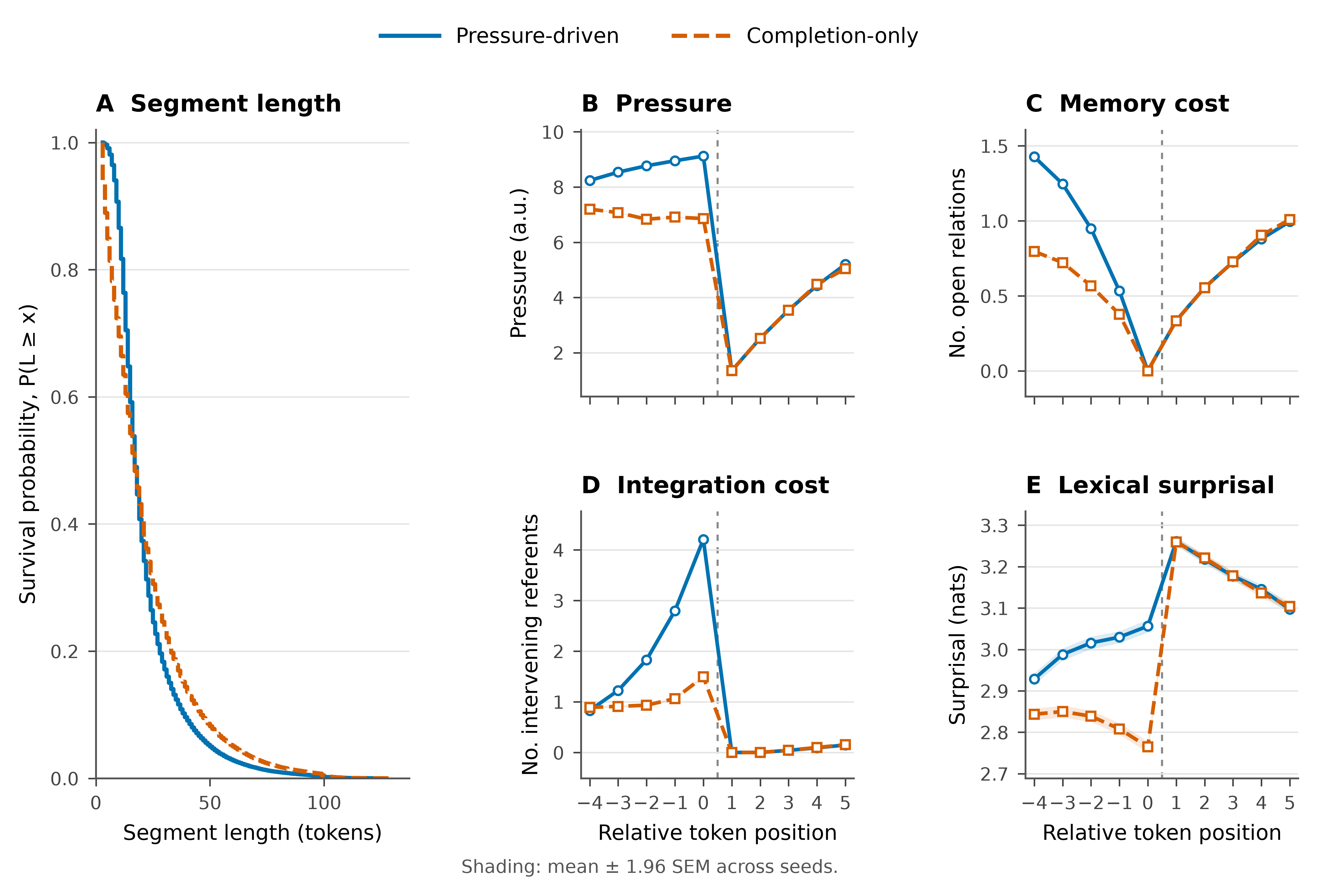}
\caption{
Pressure-dependent versus completion-only boundary formation.
(A) Survival distributions of generated segment lengths, $P(L \geq x)$, for the pressure-driven and completion-only models, calibrated to produce approximately matched mean segment lengths.
(B--E) Boundary-aligned trajectories of accumulated processing pressure (B), memory cost (C), integration cost (D), and lexical surprisal (E). Position 0 denotes the final token before boundary emission, and position +1 denotes the first token of the following segment, with the vertical dashed line marking the boundary crossing. The pressure-driven model exhibited stronger pre-boundary pressure accumulation, memory-cost reduction, and integration-cost increases, while post-boundary trajectories converged across models. Blue solid lines represent the pressure-driven model, and orange dashed lines represent the completion-only control. Shading indicates mean $\pm$ 1.96 SEM across 500 random seeds.
}
\label{fig:completion_only}
\end{figure}

To evaluate the contribution of pressure-dependent boundary gating, we compared the pressure-driven model with a completion-only control using 500 paired random seeds. The two models produced approximately matched mean segment lengths (21.57 $\pm$ 1.37 vs. 22.04 $\pm$ 1.74 tokens, mean $\pm$ standard deviation across seeds). Despite their similar mean lengths, the models differed substantially in their segment-length distributions (Figure~\ref{fig:completion_only}A). Pressure-driven segmentation showed lower length variability, with a standard deviation of 14.21 compared with 18.51 in the completion-only model ($t(499)=-33.60$, $p<0.001$). The difference persisted after normalization by mean length, with coefficients of variation of 0.657 and 0.840, respectively ($t(499)=-44.07$, $p<0.001$). The pressure-driven model also exhibited greater positive skewness (2.18 vs. 1.64, $t(499)=20.26$, $p<0.001$).

The models achieved comparable overall language-likeness scores (2.652 vs. 2.649, $p=0.432$). The completion-only model showed a slightly higher dependency-decay score (0.888 vs. 0.883, $p<0.001$), whereas the pressure-driven model showed a slightly higher Zipf score (0.769 vs. 0.760, $p=0.007$). Thus, both mechanisms reproduced the broad statistical properties included in the language-likeness criterion, despite generating different segment-length distributions.

The comparable overall language-likeness scores partly reflected offsetting differences in their component measures: the completion-only model showed slightly better dependency-length decay, while the pressure-driven model showed a better fit to Zipf-like lexical distributions. Despite similar aggregate scores, the models exhibited distinct boundary-aligned processing dynamics (Figure~\ref{fig:completion_only}B--E). In the pressure-driven model, accumulated pressure increased toward boundary emission, accompanied by a pronounced decline in memory cost and a sharp increase in integration cost immediately before closure. The completion-only model showed a weaker decline in memory cost and a substantially smaller pre-boundary integration peak. Lexical surprisal also followed different pre-boundary trajectories but converged to similar profiles after boundary emission. Both models exhibited post-boundary resets, as specified by their shared generative framework. Together, these results indicate that pressure-dependent gating contributes to the organization of segment lengths and boundary-centered processing dynamics beyond structural completion alone.

\subsection*{Robustness of processing-cost effects on boundary judgments}
\begin{figure}[ht]
\centering
\includegraphics[width=0.7\textwidth]{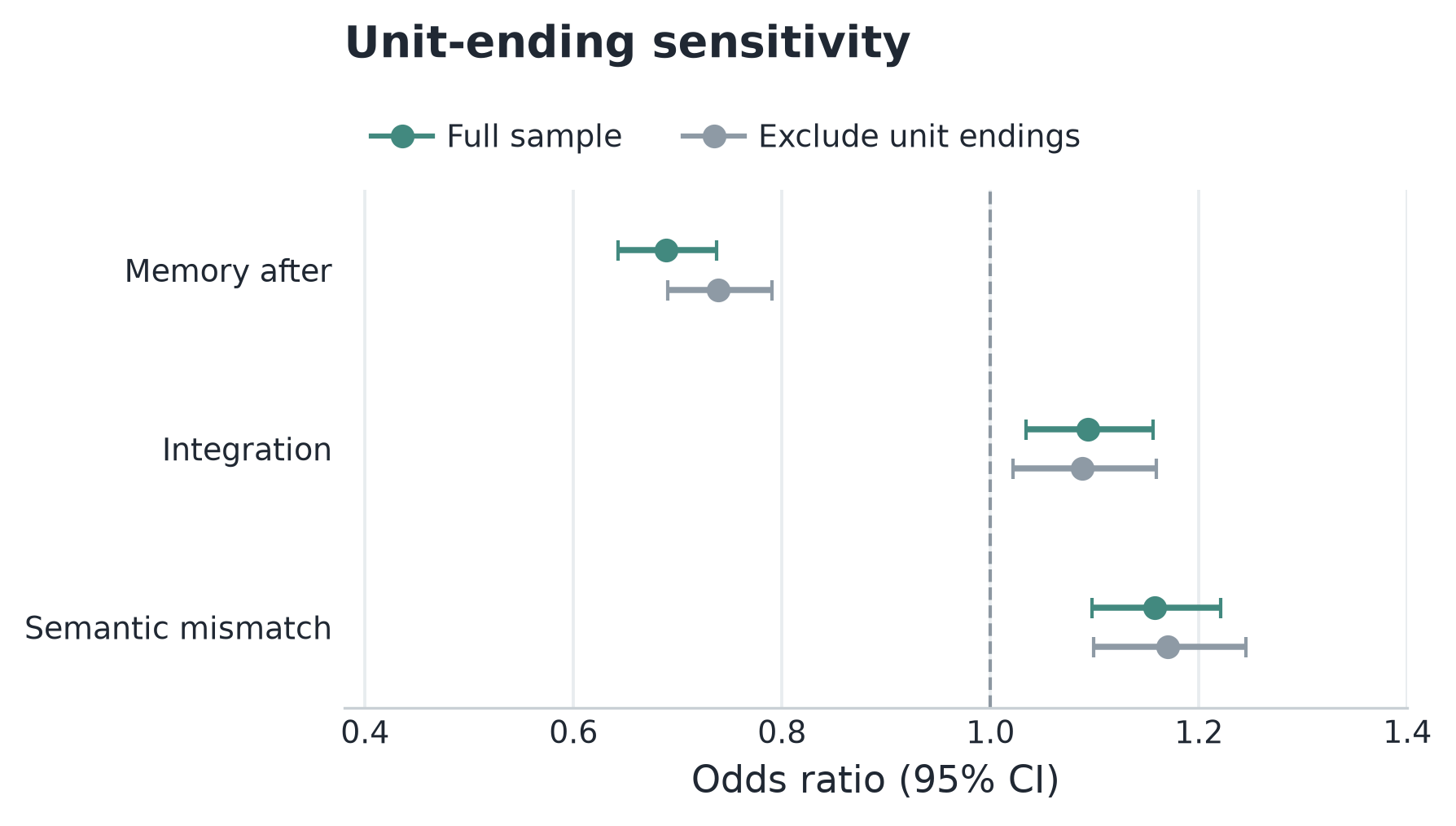}
\caption{
\textbf{Sensitivity of processing-cost effects to unit endings.}
Odds ratios and 95\% confidence intervals for memory cost after the current word, integration cost, and semantic mismatch in the full sample and after excluding 300 top-level unit-ending positions. The dashed vertical line indicates an odds ratio of 1.
}
\label{fig:chunking_si_endings}
\end{figure}

To examine whether the observed associations were driven by structurally defined unit endings, we repeated the analysis after excluding 300 annotated top-level unit-ending positions (Figure~\ref{fig:chunking_si_endings}). All three processing costs retained their directions and statistical significance: memory cost after the current word (OR = 0.739, 95\% CI [0.691, 0.791], $p<0.001$), integration cost (OR = 1.088, 95\% CI [1.022, 1.159], $p=0.008$), and semantic mismatch (OR = 1.170, 95\% CI [1.099, 1.245], $p<0.001$). Adding these predictors continued to improve model fit beyond grammatical and prosodic controls ($\Delta$AIC = $-1,101.02$; joint Wald $p<0.001$), indicating that their associations with boundary choices were not restricted to top-level unit endings.

\begin{figure}[ht]
\centering
\includegraphics[width=0.7\textwidth]{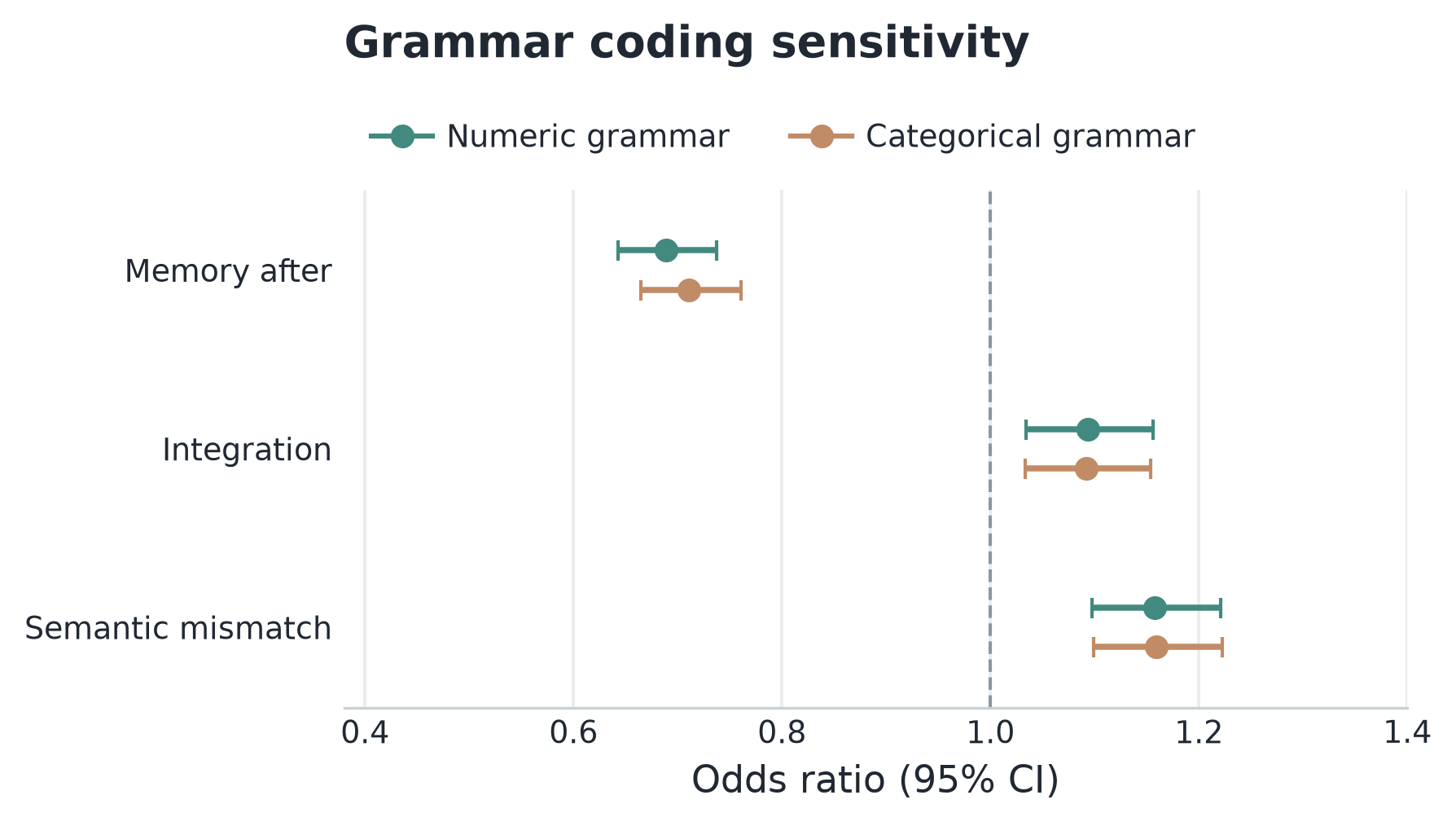}
\caption{
\textbf{Sensitivity of processing-cost effects to grammatical coding.}
Odds ratios and 95\% confidence intervals for the three processing-cost predictors under numeric and categorical representations of grammatical structure. The dashed vertical line indicates an odds ratio of 1.
}
\label{fig:chunking_si_grammar}
\end{figure}

We further tested whether these associations depended on how grammatical structure was represented. Replacing the numeric grammar predictor with a categorical specification yielded similar effects for memory cost (OR = 0.711, 95\% CI [0.665, 0.761], $p<0.001$), integration cost (OR = 1.092, 95\% CI [1.033, 1.154], $p=0.002$), and semantic mismatch (OR = 1.159, 95\% CI [1.099, 1.223], $p<0.001$; Figure~\ref{fig:chunking_si_grammar}). The three predictors also improved model fit under this alternative specification ($\Delta$AIC = $-1,362.31$; joint Wald $p<0.001$). These analyses indicate that the relationships between processing costs and perceived boundaries were robust to the exclusion of structurally defined endings and to alternative grammatical coding.

\subsection*{Boundary effect for behavioral and LLM-derived outcomes in the SPR experiment}
\begin{figure}
\centering
\includegraphics[width=\textwidth]{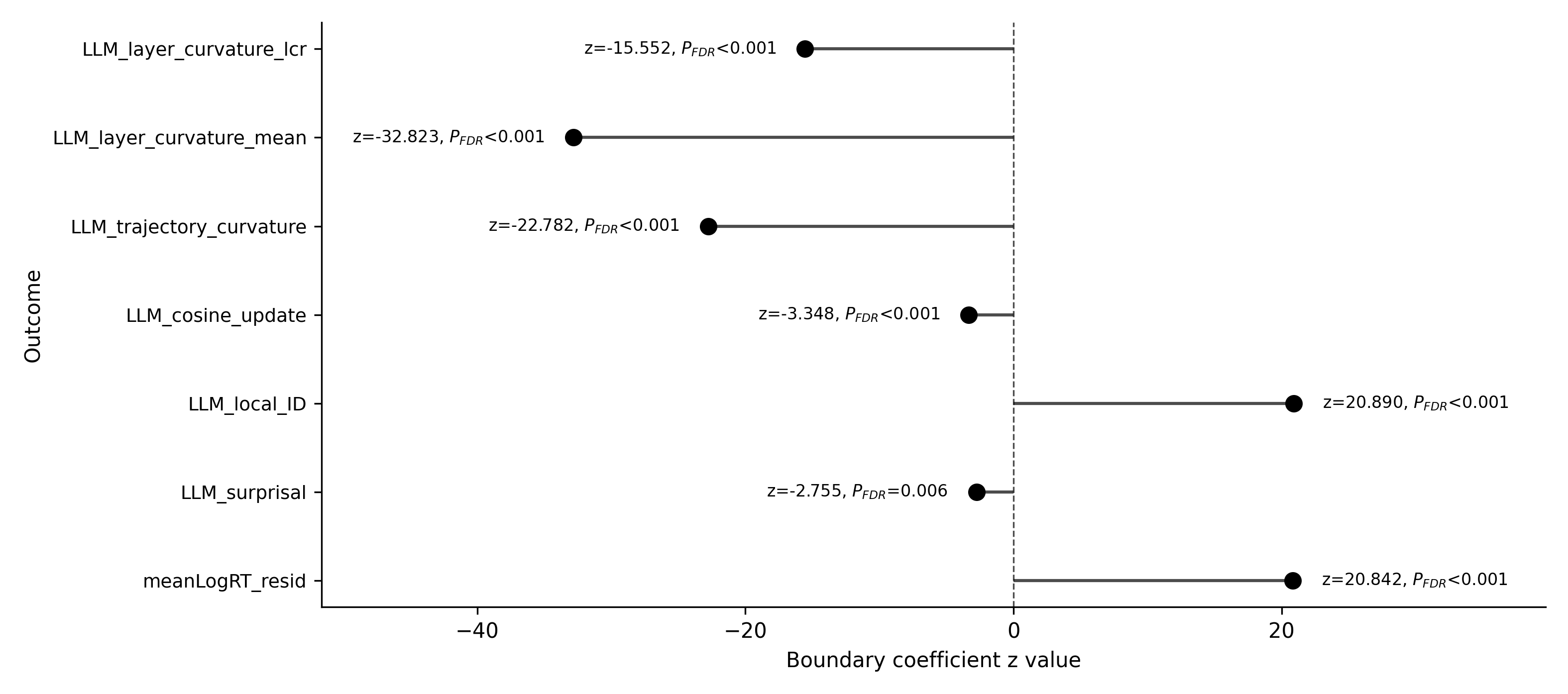}
\caption{
Sentence-boundary sensitivity differs across human behavioral and LLM representational measures.
For each outcome, the plot shows the $z$ value associated with the sentence-boundary coefficient from an ordinary least-squares regression model controlling for pressure-derived predictors, word position, and item fixed effects, with cluster-robust standard errors at the sentence level. Lines extend from zero to the estimated boundary effect. Positive values indicate increased outcome values at sentence boundaries, whereas negative values indicate decreased outcome values at sentence boundaries. Reading-time residuals and LLM local intrinsic dimensionality increased at boundaries, while LLM surprisal, cosine update, trajectory curvature, and layer-curvature measures decreased at boundaries. Reported $P_{\mathrm{FDR}}$ values are corrected across outcomes using False Discovery Rate (FDR).
}
\label{fig:boundary_coeff}
\end{figure}

After accounting for pressure-derived predictors, word position, and item fixed effects, explicit sentence-boundary status still showed statistically significant associations with all behavioral and LLM-derived outcomes (Figure~\ref{fig:boundary_coeff}). 

\vspace{0.4em}

\begin{table}[ht]
\centering
\caption{Fixed architectural parameters used in the parameter scan.}
\label{tab:fixed_parameters}
\begin{tabular}{lll}
\hline
Parameter & Value & Interpretation \\
\hline
\(k\) & \(1.2\) & Steepness of the pressure-to-boundary sigmoid \\
\(N^{\mathrm{open}}_{\max}\) & \(6\) & Maximum number of simultaneously open relations \\
\(L_{\min}\) & \(3\) & Minimum segment length \\
\(L_{\max}\) & \(99\) & Maximum segment length \\
\(P_{\mathrm{reset}}\) & \(0.0\) & Pressure value after a boundary reset \\
\(\sigma_{\epsilon}\) & \(0.05\) & Standard deviation of pressure noise \\
\(V\) & \(200\) & Vocabulary size \\
\(d\) & \(32\) & Word embedding dimensionality \\
\(K\) & \(8\) & Number of latent semantic topics \\
\(\sigma_{\mathrm{emb}}\) & \(0.25\) & Noise level in synthetic word embeddings \\
\(\eta\) & \(0.15\) & Semantic context update rate \\
Lexical seed & \(42\) & Fixed seed for constructing the lexical space \\
\hline
\end{tabular}
\end{table}

\vspace{0.4em}

\begin{table}[ht]
\caption{Tuned parameters used in the parameter scan.}
\label{tab:tuned_parameters}

\begin{tabular}{@{}p{0.16\textwidth}p{0.32\textwidth}p{0.34\textwidth}p{0.13\textwidth}@{}}
\hline
Parameter & Candidate values & Interpretation & Applied family \\
\hline

\(\theta\)
& 5.0, 8.0, 12.0
& Boundary threshold; controls how much pressure is required before boundary drive increases.
& All \\

\(\rho\)
& 0.3, 0.5, 0.8, 0.95
& Pressure carry-over; controls how strongly pressure persists across events.
& All \\

\((b_{\mathrm{open}}, b_{\mathrm{close}})\)
& (0.15, 0.15), (0.30, 0.30), (0.50, 0.50), (0.15, 0.50), (0.50, 0.15)
& Baseline tendency to open and close dependency-like relations.
& All \\

\((a_{\mathrm{mem}}, b_{\mathrm{int}})\)
& (0.2, 0.0), (0.0, 0.2), (0.2, 0.2), (0.5, 0.2), (0.2, 0.5), (0.5, 0.5)
& DLT-inspired memory and integration cost weights.
& DLT-only; DLT-surprisal \\

\(\gamma_{\mathrm{surp}}\)
& 0.0, 0.02, 0.05, 0.1, 0.3, 0.6, 0.9
& Weight of lexical surprisal in the pressure equation.
& Surprisal-only \\

\(\gamma_{\mathrm{surp}}\)
& 0.02, 0.05, 0.1, 0.3, 0.6, 0.9
& Non-zero lexical surprisal weight; ensures that lexical surprisal is active.
& DLT-surprisal \\

\(\beta_{\mathrm{ctx}}\)
& 0.0, 1.0, 3.0, 6.0, 9.0, 12.0, 15.0
& Semantic context strength; controls how strongly context shapes lexical sampling.
& All \\

\hline
\end{tabular}
\end{table}

\begin{table}[ht]
\centering
\caption{The best parameter regime.}
\label{tab:best_parameter_regime}
\begin{tabular}{lll}
\hline
Parameter & Value & Interpretation \\
\hline
Model family & DLT--surprisal & Combined structural and contextual pressure model \\
Combination ID & 13777 & Parameter combination selected by language-likeness score \\
$b_{\mathrm{open}}$ & 0.50 & Baseline tendency to open dependency-like relations \\
$b_{\mathrm{close}}$ & 0.50 & Baseline tendency to close dependency-like relations \\
$a_{\mathrm{mem}}$ & 0.50 & Weight of memory cost from currently open relations \\
$b_{\mathrm{int}}$ & 0.20 & Weight of integration cost from resolved relations \\
$\gamma_{\mathrm{surp}}$ & 0.30 & Weight of contextual surprisal in processing pressure \\
$\beta_{\mathrm{ctx}}$ & 9.00 & Strength of contextual constraint on lexical sampling \\
$\theta$ & 8.00 & Boundary threshold \\
$\rho$ & 0.80 & Pressure carry-over across time steps \\
Context condition & context-sensitive & Lexical probabilities depended on semantic context \\
Mean language-likeness score & 2.680 & Mean score used for regime selection \\
Quality-control pass rate & 1.00 & Proportion of simulations passing anti-collapse criteria \\
\hline
\end{tabular}
\end{table}
\clearpage

\clearpage

\clearpage

\clearpage

\clearpage

\clearpage

\end{document}